%% file: ms.tex
\documentclass[11pt,a4paper]{article}
\usepackage[hyperref]{acl2020}
\usepackage{times}
\usepackage{latexsym}

\usepackage{caption}
\usepackage{subcaption}
\usepackage{graphicx}
\usepackage{booktabs}

\usepackage{amsmath}
\usepackage{adjustbox}

\usepackage{multirow}
\usepackage{longtable}

\usepackage[utf8]{inputenc}

\usepackage{microtype}

\usepackage{array}
\usepackage{xspace}

\usepackage{xcolor,colortbl}
\usepackage{tikz}
\usepackage{collcell}
\usepackage{xstring}
\usepackage{textcomp}

\newcommand*{\MinNumber}{1}%
\newcommand*{\MidNumber}{60} %
\newcommand*{\MaxNumber}{983}%

\definecolor{darkblueer}{HTML}{5E35B1}
\definecolor{midblueer}{HTML}{9575CD}
\definecolor{lightblueer}{HTML}{EDE7F6}

\definecolor{darkblueed}{HTML}{1E88E5}
\definecolor{midblueed}{HTML}{90CAF9}
\definecolor{lightblueed}{HTML}{E3F2FD}

\def\zz#1{
    \ifdim#1pt<\MidNumber pt
        \pgfmathsetmacro{\PercentColor}{100*((\MidNumber - #1)/(\MidNumber - \MinNumber))}
        \xdef\PercentColorr{\PercentColor}
        \cellcolor{lightblueed!\PercentColorr!midblueed!60}{#1}
    \else
        \pgfmathsetmacro{\PercentColor}{100*((#1 - \MidNumber)/(\MaxNumber - \MidNumber))} %
        \xdef\PercentColorr{\PercentColor}
        \cellcolor{darkblueed!\PercentColorr!midblueed!60}{#1}
    \fi
}

\newcommand*{\MinMRR}{0.38}%
\newcommand*{\MidMRR}{0.55} %
\newcommand*{\MaxMRR}{0.73}%

\def\mrr#1{
    \ifdim#1pt<\MidMRR pt
        \pgfmathsetmacro{\PercentColor}{100*((\MidMRR - #1)/(\MidMRR - \MinMRR))}
        \xdef\PercentColorr{\PercentColor}
        \cellcolor{lightblueer!\PercentColorr!midblueer!60}{#1}
    \else
        \pgfmathsetmacro{\PercentColor}{100*((#1 - \MidMRR)/(\MaxMRR - \MidMRR))} %
        \xdef\PercentColorr{\PercentColor}
        \cellcolor{darkblueer!\PercentColorr!midblueer!60}{#1}
    \fi
}

\newcommand{\vbl}{{\vrule width 1.1pt}}
\newcommand{\hbl}{\noalign{
\hrule height 1.1pt
}}
\newcommand{\pad}{1.3}

\newcolumntype{P}[1]{>{\centering\arraybackslash}p{#1}}

\newcommand{\custo}{\bf \fontfamily{bch}\selectfont}

\newcommand{\confacl}{\mbox{{\custo ACL}}\xspace}
\newcommand{\confcl}{\mbox{{\custo CL}}\xspace}
\newcommand{\confcoling}{\mbox{{\custo COLING}}\xspace}
\newcommand{\confconll}{\mbox{{\custo CONLL}}\xspace}
\newcommand{\confeacl}{\mbox{{\custo EACL}}\xspace}
\newcommand{\confemnlp}{\mbox{{\custo EMNLP}}\xspace}
\newcommand{\conflrec}{\mbox{{\custo LREC}}\xspace}
\newcommand{\confnaacl}{\mbox{{\custo NAACL}}\xspace}
\newcommand{\confsemeval}{\mbox{{\custo SEMEVAL}}\xspace}
\newcommand{\conftacl}{\mbox{{\custo TACL}}\xspace}
\newcommand{\confws}{\mbox{{\custo WS}}\xspace}

\newcommand{\LANG}[1]{{\fontfamily{lmtt}\selectfont #1}}
\aclfinalcopy 
\def\aclpaperid{1751} 

\title{The State and Fate of Linguistic Diversity and Inclusion in the NLP World}

\author{ \textbf{Pratik Joshi\thanks{\hspace{0.1cm} Authors contributed equally to the work.}}\quad \textbf{Sebastin Santy\footnotemark[1]} \quad \textbf{Amar Budhiraja\footnotemark[1]}  \\\textbf{Kalika Bali} \quad  \textbf{Monojit Choudhury} \\
Microsoft Research, India \\
\{t-prjos, t-sesan, amar.budhiraja, kalikab, monojitc\}@microsoft.com
}

\date{}

\newcommand\blfootnote[1]{%
  \begingroup
  \renewcommand\thefootnote{}\footnote{#1}%
  \addtocounter{footnote}{-1}%
  \endgroup
}

\begin{document}
\maketitle
\begin{abstract}
 Language technologies contribute to promoting multilingualism and linguistic diversity around the world. However, only a very small number of the over 7000 languages of the world are represented in the rapidly evolving language technologies and applications. In this paper we look at the relation between the types of languages, resources, and their representation in NLP conferences to understand the trajectory that different languages have followed over time. Our quantitative investigation underlines the disparity between languages, especially in terms of their resources, and calls into question the ``language agnostic" status of current models and systems. Through this paper, we attempt to convince the ACL community to prioritise the resolution of the predicaments highlighted here, so that no language is left behind.
\end{abstract}

\input{sections/1_introduction.tex}
\input{sections/2_taxonomy.tex}
\input{sections/3_typology.tex}
\input{sections/4_analysis.tex}

\input{sections/5_embeddings_v3.tex}
\input{sections/6_discussion.tex}
\input{sections/7_conclusion.tex}
\bibliography{acl2020}
\bibliographystyle{acl_natbib}
\appendix
\input{sections/8_appendix}

\end{document}

%% file: sections/1_introduction.tex
\section{The Questions}
\begin{figure}
    \centering
    \begin{subfigure}{0.24\textwidth}
            \includegraphics[width=\columnwidth,keepaspectratio]{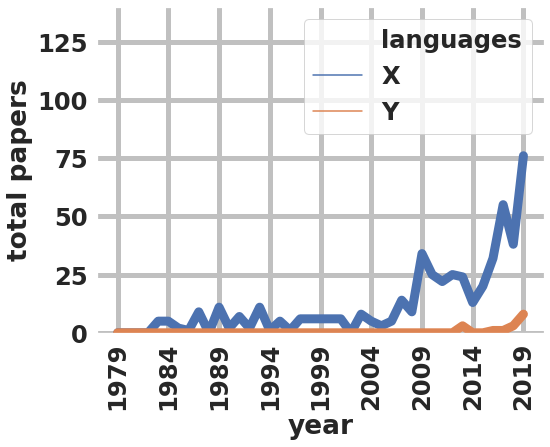}
            \caption{\scriptsize \confacl + \confnaacl + \confeacl + \confemnlp}
    \end{subfigure}\hfill
    \begin{subfigure}{0.24\textwidth}
            \includegraphics[width=\columnwidth,keepaspectratio]{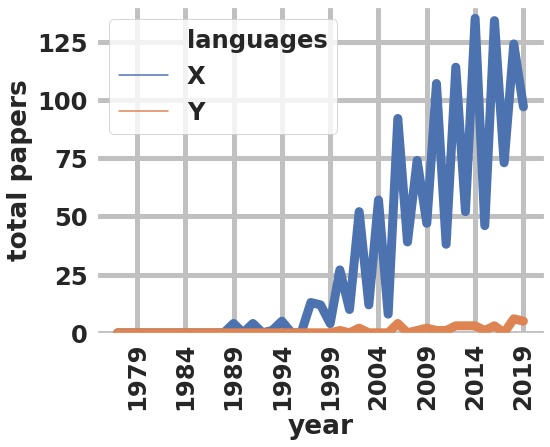}
            \caption{\scriptsize \conflrec + \confws}
    \end{subfigure}
    \caption{Number of papers with mentions of {\bf X} and {\bf Y} language for two sets of conferences.}
    \label{fig:introblah}
\end{figure}

\blfootnote{\href{https://microsoft.github.io/linguisticdiversity}{https://microsoft.github.io/linguisticdiversity}}
Languages {\bf X} and {\bf Y} are the official languages of two different countries; they have around 29M and 18M native speakers, and 2M and 5.5K Wikipedia articles, respectively. {\bf X} is syntactically quite similar to English, though uses dimunitives and has grammatical gender. {\bf Y}, on the other hand, has a different word order from English, and has a rare typological feature - generally it is a head-final language, but noun phrases are head-initial. It also features full and partial reduplication. 69 items on LDC and ELRA contain data in {\bf X}, whereas for {\bf Y} there are only 2 items.  {\bf X} boasts of some of the best online machine translation systems, whereas {\bf Y} is supported by very few online MT systems and that too with far inferior translation quality.  Figure \ref{fig:introblah} shows the number of papers in conferences (\confacl, \confnaacl, \confeacl, \confemnlp, \conflrec, \confws) that mention {\bf X} and {\bf Y} in the paper, across the years. As you can see, while {\bf X} has a steady and growing trend of research, our community has been mostly oblivious to {\bf Y}, until recently when some of the zero-shot learning papers have started mentioning it.  Can you guess what {\bf X} and {\bf Y} are?

Regardless of whether you can guess the exact answer, most NLP researchers surely know of (and might even speak) several languages which are in the same boat as {\bf X}; languages which have a large amount of resources and therefore access to the benefits of the current NLP breakthroughs, and languages like {\bf Y}; those which lack resources and consequently the attention of the NLP community, despite having similar speaker base sizes and typologically diverse features.


You probably have come across the issue of extremely skewed distribution of resources across the world's languages before. You might also be aware of the fact that most of our NLP systems, which are typically declared language agnostic, are not truly so \cite{Bender2011OnAA}. The handful of languages on which NLP systems are trained and tested are often related and from the same geography, drawn from a few dominant language families, leading to a typological echo-chamber. As a result, a vast majority of typologically diverse linguistic phenomena are never seen by our NLP systems \cite{typologicalsurvey}. Nevertheless, it would be prudent to re-examine these issues in the light of recent advances in deep learning. Neural systems, on one hand, require a lot more data for training than rule-based or traditional ML systems, creating a bigger technological divide between the {\bf X}s and {\bf Y}s; yet, some of the most recent techniques on zero-shot learning of massively multilingual systems \cite{bertdevlin, conneau2019cross, aharoni-etal-2019-massively, massivemultilingualholger} bridge this gap by obliterating the need for large labeled datasets in all languages. Instead, they need only large unlabeled corpora across languages and labeled data in only some languages. Assuming that this approach can be taken to its promising end, how does the fate of different languages change?

We break down this complex prescient question into the following more tractable and quantifiable questions on {\em Linguistic Diversity and Inclusion}:
\vskip 0.1cm
\noindent
\textbf{1.} How many resources, labeled and unlabeled, are available across the World's languages? How does this distribution correlate to their number of native speakers? What can we expect to achieve today and in the near future for these languages?
\vskip 0.1cm
\noindent
\textbf{2.} Which typological features have current NLP systems been exposed to, and which typological features mostly remain unexplored by systems because we have hardly created any resources and conducted data-driven research in those languages?
\vskip 0.1cm
\noindent
\textbf{3.} As a community, how inclusive has \confacl been in conducting and publishing research on various languages? In 1980s and early 90s, when large scale datasets were not the prime drivers of research, was the linguistic diversity of \confacl higher than what it has been in 2000s and 2010s? Or has \confacl become more inclusive and diverse over the years?
\vskip 0.1cm
\noindent
\textbf{4.} Does the amount of resource available in a language influence the research questions and the venue of publication? If so, how?
\vskip 0.1cm
\noindent
\textbf{5.} What role does an individual researcher, or a research community have to play in bridging the linguistic-resource divide?
\vskip 0.1cm
\noindent

In this paper, we take a multi-pronged quantitative approach to study and answer the aforementioned questions, presented in order, in the following five sections. One of the key findings of our study, to spill the beans a bit, is that the languages of the World can be broadly classified into 6 classes based on how much and what kind of resources they have; the languages in each class have followed a distinct and different trajectory in the history of \confacl, and some of the hitherto neglected classes of languages have more hope of coming to the forefront of NLP technology with the promised potential of zero-shot learning. 

%% file: sections/2_taxonomy.tex
\section{The Six Kinds of Languages}

In order to summarize the digital status and `richness' of languages in the context of data availability, we propose a taxonomy based on the number of language resources which exist for different languages. We frame the rest of our analyses based on this taxonomy and use it to emphasize the existence of such resource disparities.

\subsection{Features}
We design this taxonomy using two feature axes: number of unlabeled resources vs. number of labeled resources. Previous methods have mostly relied on supervised learning techniques which require labeled corpora. However, the advent of transfer learning methods have boosted the importance of unlabeled data: massively multilingual models such as mBERT use Wikipedia for pre-training, and then fine-tune on downstream NLP tasks. These features are suitable because the current NLP research is predominantly data-driven, and language inclusion depends on how much labeled or unlabeled data is available. We believe these features are sufficient for the taxonomical design as the required metadata is consistently available across all languages, whereas features such as number of hours required to collect data aren't available.



We treat each data resource as a fundamental unit, based on the assumption that the collection of one unit is proportional to a certain extent of effort being invested towards the resource improvement of that language. Moreover, this feature discretization is unambiguous and concrete. Other units such as the total number of datapoints across datasets can be misleading because different NLP tasks have different data requirements. For example, while Machine Translation (MT) models require datapoints to the order of millions \cite{koehn2017six} to perform competitively, competent models in Question Answering require around 100 thousand datapoints \cite{rajpurkar2016squad}. Moreover, the unit of datapoints vary across different technologies (e.g. Speech data measured in hours, MT data measured in number of parallel sentences).


\subsection{Repositories}
We focus our attention on the LDC catalog\footnote{\href{https://catalog.ldc.upenn.edu/}{https://catalog.ldc.upenn.edu/}} and the ELRA Map\footnote{\href{http://catalog.elra.info/en-us/}{http://catalog.elra.info/en-us/}} for labeled datasets. Although there are other repositories of data available online, we found it practical to treat these organized collections as a representation of labeled dataset availability. This way, we look at standardized datasets that have established data quality and consistency, and which have been used in prior work. There are strong efforts such as PanLex \cite{panlex}, which is a large lexical database of a wide range of languages being used for a lexical translator, and OLAC \cite{olac}, which contains a range of information for different languages (e.g. text collections, audio recordings, and dictionaries). However, keeping within the purview of NLP datasets used in *CL conferences, we decided to focus on popular repositories such as the above-mentioned.

We look at Wikipedia pages as a measure for unlabeled data resources. With regards to language technologies, Wikipedia pages represent a strong source of unsupervised training data which are freely and easily accessible. In the perspective of digital resource availability, they are a comprehensive source of factual information and are accessed by a large, diverse set of online users.

\begin{figure}[!t]
    \centering
    \includegraphics[width=\linewidth,keepaspectratio]{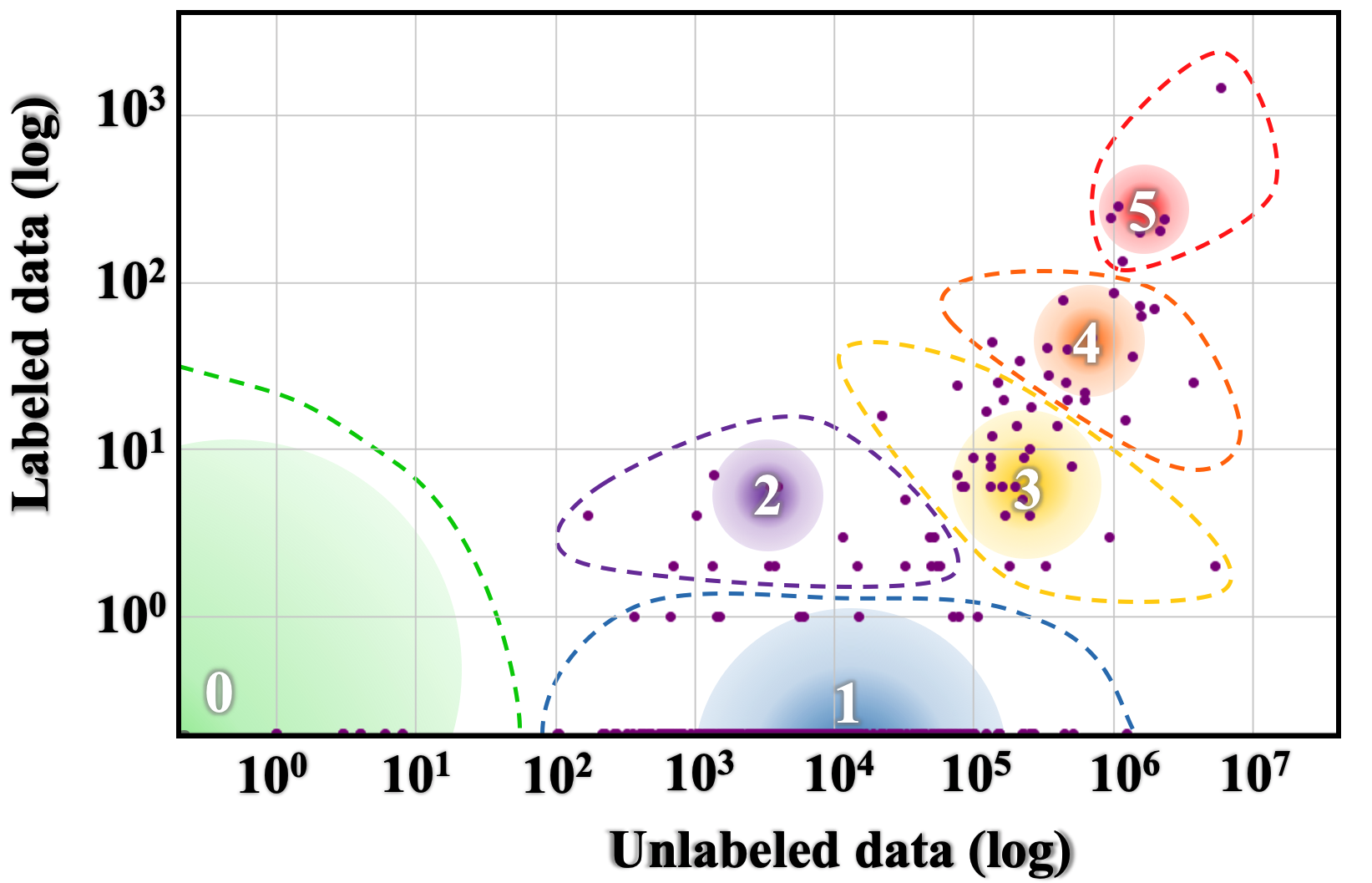}
    
    \caption{Language Resource Distribution: The size of the gradient circle represents the number of languages in the class. The color spectrum VIBGYOR, represents the total speaker population size from low to high. Bounding curves used to demonstrate covered points by that language class.}
    \label{fig:taxonomy}
\end{figure}


\begin{table*}[!t]
\begin{center}
\small
{\renewcommand{\arraystretch}{\pad}
\begin{tabular}{! \vbl >{\columncolor[HTML]{EFEFEF}}c|c|c|c|c! \vbl}

      \hbl
      \cellcolor[HTML]{C0C0C0}\textbf{Class} & \cellcolor[HTML]{EFEFEF}\textbf{5 Example Languages} & \cellcolor[HTML]{EFEFEF}\textbf{\#Langs} & \cellcolor[HTML]{EFEFEF}\textbf{\#Speakers} & \cellcolor[HTML]{EFEFEF}\textbf{\% of Total Langs} \\
      \hline
     0 & \LANG{Dahalo}, \LANG{Warlpiri}, \LANG{Popoloca}, \LANG{Wallisian}, \LANG{Bora} & 2191 & 1.0B & 88.17\% \\
      \hline
     1 & \LANG{Cherokee}, \LANG{Fijian}, \LANG{Greenlandic}, \LANG{Bhojpuri}, \LANG{Navajo} & 222 & 1.0B & 8.93\% \\
      \hline
     2 & \LANG{Zulu}, \LANG{Konkani}, \LANG{Lao}, \LANG{Maltese}, \LANG{Irish} & 19 & 300M & 0.76\% \\
      \hline
     3 & \LANG{Indonesian}, \LANG{Ukranian}, \LANG{Cebuano}, \LANG{Afrikaans}, \LANG{Hebrew} & 28 & 1.1B & 1.13\% \\
      \hline
     4 & \LANG{Russian}, \LANG{Hungarian}, \LANG{Vietnamese}, \LANG{Dutch}, \LANG{Korean} & 18 & 1.6B & 0.72\% \\
      \hline
     5 & \LANG{English}, \LANG{Spanish}, \LANG{German}, \LANG{Japanese}, \LANG{French} & 7 & 2.5B & 0.28\% \\
      \hbl

\end{tabular}}
\caption{Number of languages, number of speakers, and percentage of total languages for each language class.}
\label{tab:taxstats}
 \end{center}
\end{table*}

\begin{figure*}[!t]
    \centering
        \begin{subfigure}{0.24\textwidth}
            \includegraphics[width=\columnwidth,keepaspectratio]{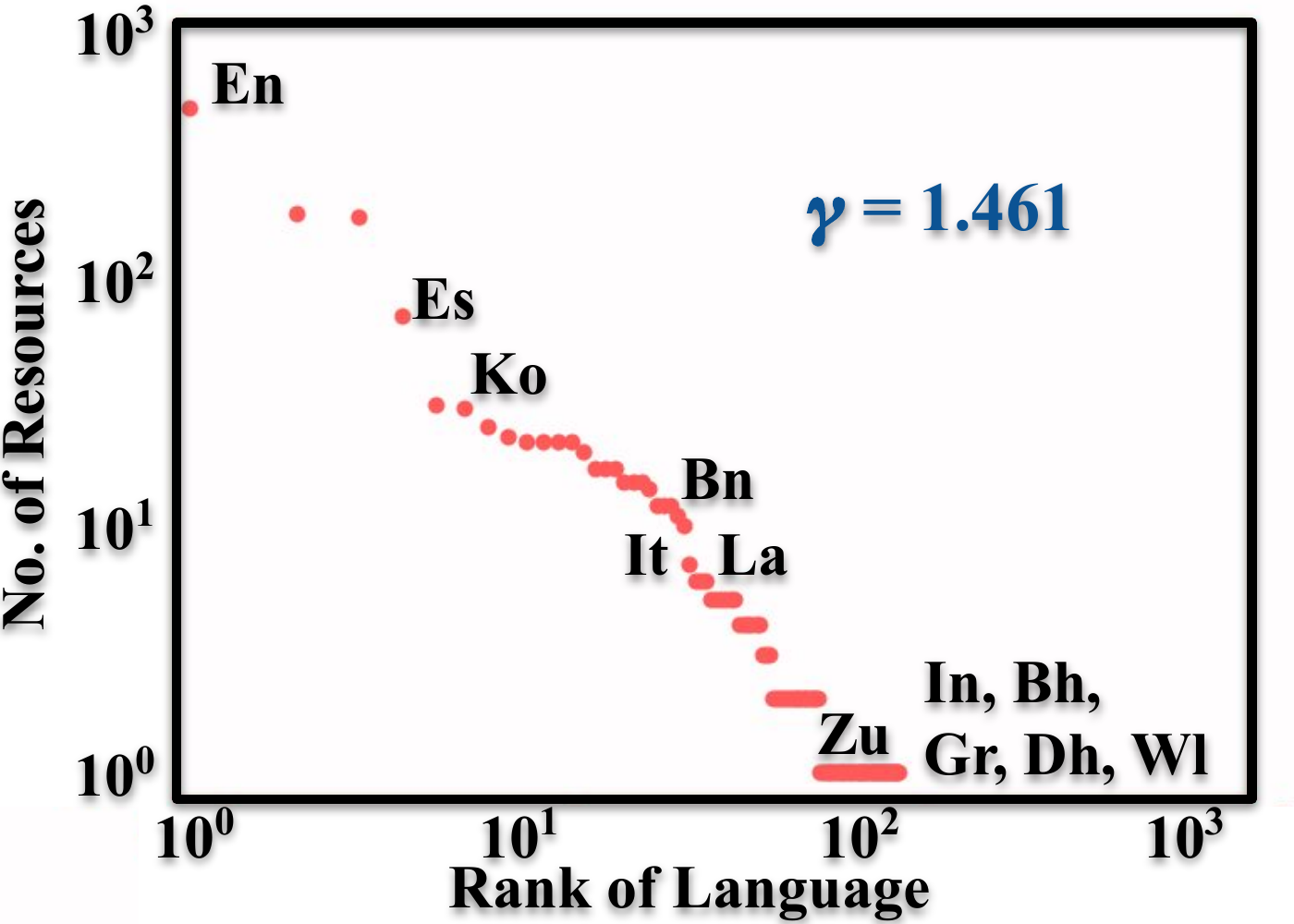}
            \caption{LDC}
        \end{subfigure} \hfill
        \begin{subfigure}{0.24\textwidth}
            \includegraphics[width=\columnwidth,keepaspectratio]{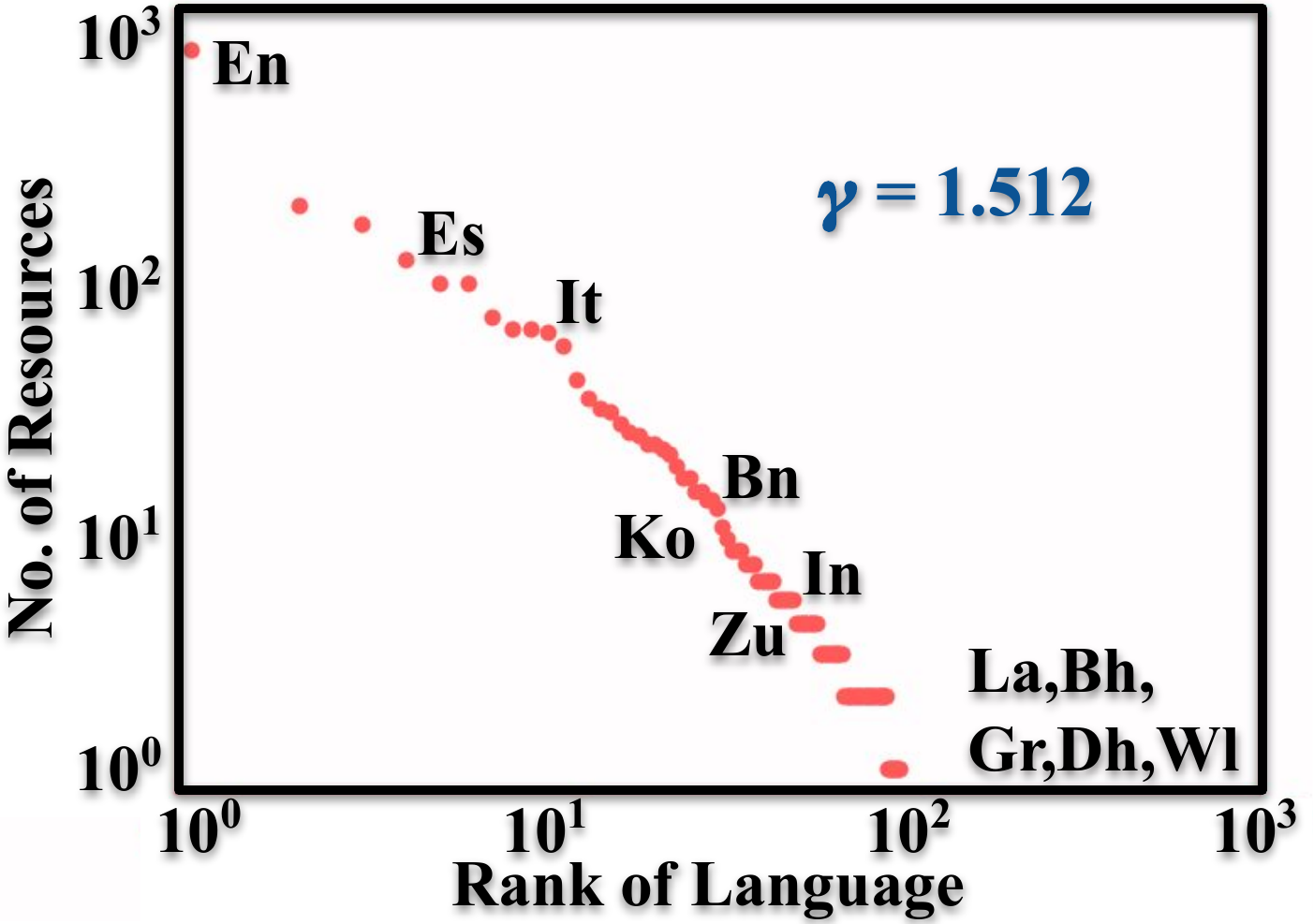}
            \caption{LRE}
        \end{subfigure} \hfill
        \begin{subfigure}{0.24\textwidth}
            \includegraphics[width=\columnwidth,keepaspectratio]{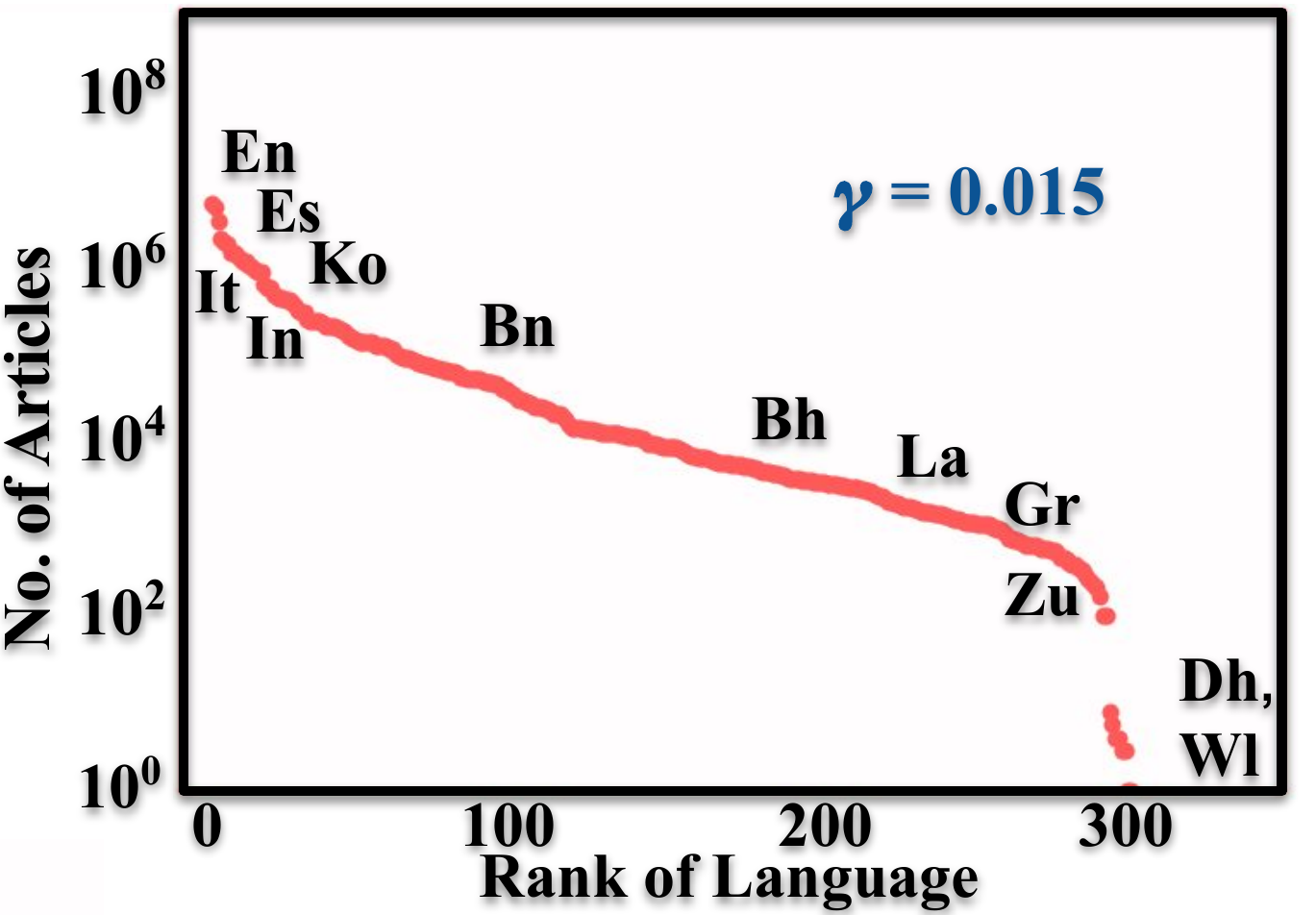}
            \caption{Wikipedia}
        \end{subfigure} \hfill
        \begin{subfigure}{0.24\textwidth}
            \includegraphics[width=\columnwidth,keepaspectratio]{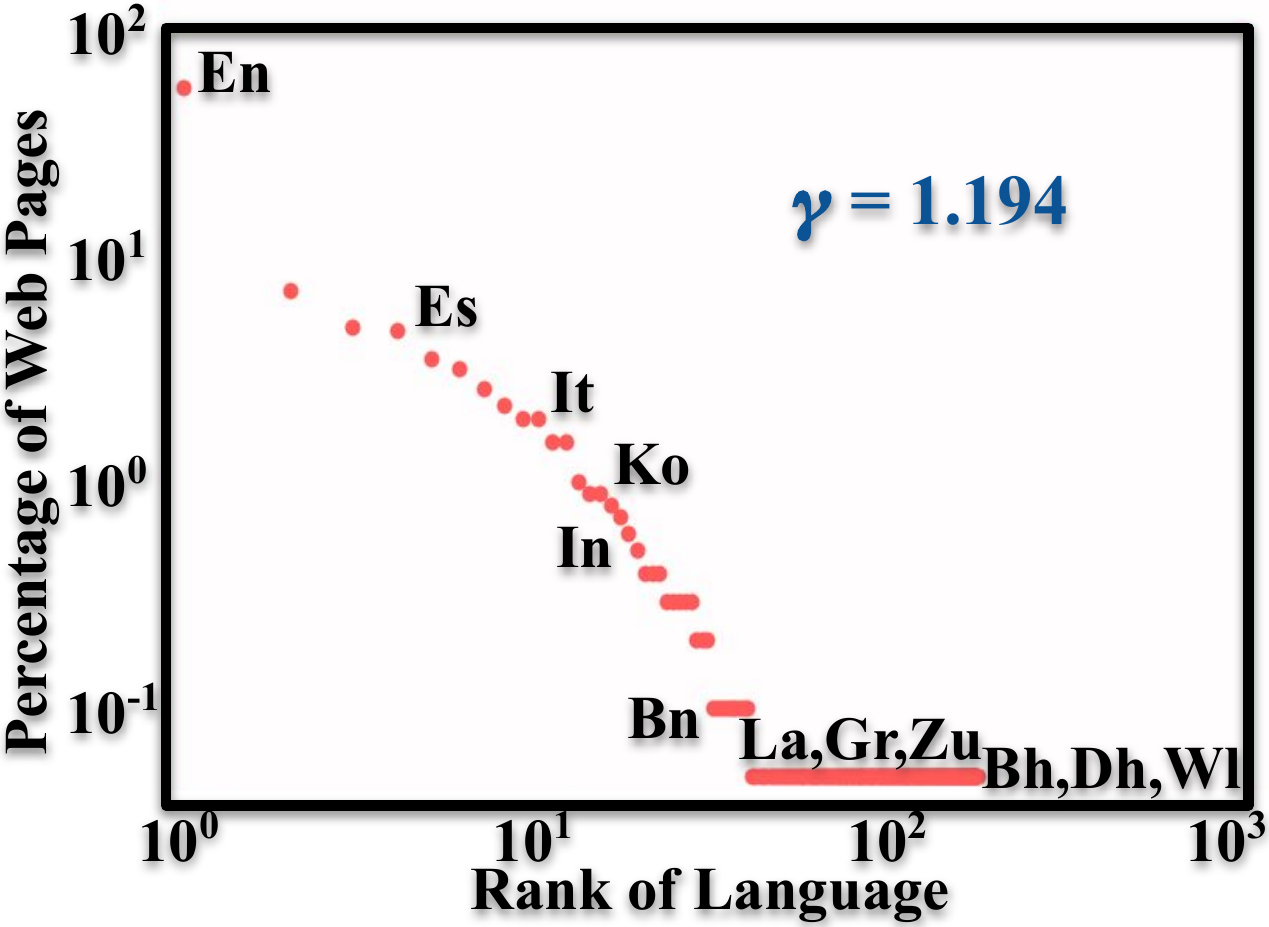}
            \caption{Web}
        \end{subfigure}
        \caption{Plots of different available resources for different languages. Languages to the far right do not have a representation in the resource category. Languages annotated are: Class 0-\LANG{Dahalo} (Dh), \LANG{Wallisian}(Wl); Class 1-\LANG{Bhojpuri} (Bh), \LANG{Greenlandic} (Gr); Class 2-\LANG{Lao} (La), \LANG{Zulu} (Zu); Class 3- \LANG{Bengali} (Bn), \LANG{Indonesian} (In); Class 4- \LANG{Korean} (Ko), \LANG{Italian} (It); Class 5- \LANG{English} (En), \LANG{Spanish} (Es).}
        \label{fig:powerlaw}
\end{figure*}
\subsection{Language Classes}
Figure \ref{fig:taxonomy} is a visualization of the taxonomy. We find a set of distinct partitions which can be used to categorize languages into 6 unique positions in the language resource `race':
\vskip 0.15cm
\noindent
    \textbf{0 - The Left-Behinds} These languages have been and are still ignored in the aspect of language technologies. With exceptionally limited resources, it will be a monumentous, probably impossible effort to lift them up in the digital space. Unsupervised pre-training methods only make the `poor poorer', since there is virtually no unlabeled data to use.
    \vskip 0.15cm
\noindent
    \textbf{1 - The Scraping-Bys} With some amount of unlabeled data, there is a possibility that they could be in a better position in the `race' in a matter of years. However, this task will take a solid, organized movement that increases awareness about these languages, and also sparks a strong effort to collect labelled datasets for them, seeing as they have almost none.
    \vskip 0.15cm
\noindent
    \textbf{2 - The Hopefuls} With light at the end of the tunnel, these languages still fight on with their gasping breath. A small set of labeled datasets has been collected for these languages, meaning that there are researchers and language support communities which strive to keep them alive in the digital world. Promising NLP tools can be created for these languages a few years down the line.
    \vskip 0.15cm
\noindent
    \textbf{3 - The Rising Stars} Unsupervised pre-training has been an energy boost for these languages. With a strong web presence, there is a thriving cultural community online for them. However, they have been let down by insufficient efforts in labeled data collection. With the right steps, these languages can be very well off if they continue to ride the `pre-training' wave.
    \vskip 0.15cm
\noindent
    \textbf{4 - The Underdogs} Powerful and capable, these languages pack serious amounts of resource `firepower'. They have a large amount of unlabeled data, comparable to those possessed by the winners, and are only challenged by lesser amount of labeled data. With dedicated NLP communities conducting research on these languages, they have the potential to become winners and enjoy the fruits of `digital superiority'. 
    \vskip 0.15cm
\noindent
    \textbf{5 - The Winners} Running strong and fast, these languages have been in the lead for quite a while now, some longer than others. With a dominant online presence, there have been massive industrial and government investments in the development of resources and technologies for these languages. They are the quintessential rich-resource languages, reaping benefit from each state-of-the-art NLP breakthrough. 

Some more information about the taxonomy is shown in Table \ref{tab:taxstats}. We also take 10 languages, and annotate their positions in Figure \ref{fig:powerlaw}.




\subsection{Findings}
\vskip 0.1cm
\textbf{On your marks} As can be seen in Figure \ref{fig:powerlaw}, the Winners take pole position in all rankings, and Class 0 languages remain `out of the race' with no representation in any resource. The Wikipedia distribution seems to be more fair for classes 1, 2, and 3 when compared to classes 4 and 5, whereas the Web distribution has a clear disparity.\vskip 0.2cm
\noindent
\textbf{Talk ain't cheap} Looking at Table \ref{tab:taxstats}, we see that Class 0 contains the largest section of languages and represents 15\% of all speakers across classes. Although there is a large chunk of speakers which converse with Class 5 languages, the lack of technological inclusion for different languages could draw native speakers away from Class 0 languages and towards Class 5, exacerbating the disparity.

%% file: sections/3_typology.tex
\section{Typology}

Linguistic typology is a field which involves the classification of languages based on their structural and semantic properties. Large-scale efforts have led to the creation of a database of typological features \cite{wals}. Such documentation becomes important as there are barely any other classifications of similar scale. In the context of NLP research, there has been work indicating the effectiveness of injecting typological information to guide the design of models \cite{typologicalsurvey}. Also, transfer learning of resource-rich to resource-poor languages have been shown to work better if the respective languages contain similar typological features \cite{pireshowmultilingual}. We look at how skewed language resource availability leads to an under-representation of certain typological features, which may in turn cause zero-shot inference models to fail on NLP tasks for certain languages. 


We look at the WALS data \cite{wals}, which contains typological features for 2679 languages. There are a total of 192 typological features, with an average of 5.93 categories per feature. We take the languages in classes 0, 1, 2, all of which have limited or no data resources as compared to 3, 4, 5 and look at how many categories, across all features, exist in classes 0, 1, 2 but not 3, 4, 5. This comes to a total of 549 out of 1139 unique categories, with an average of 2.86 categories per feature being ignored. Typological features with the most and least `ignored' categories are shown in Table \ref{tab:features}. 

To get an idea of what these typological `exclusions' mean in the context of modern multilingual methods, we look at the specific languages which contain these `excluded' categories in the respective features, and compare their performances in similarity search, from the results of \citet{massivemultilingualholger}. Table \ref{tab:typology} shows some examples of how `ignored' features have been difficult to deal with even when jointly training of all languages.

\begin{table}[!t]
\begin{center}
\footnotesize
{\renewcommand{\arraystretch}{\pad}
\begin{tabular}{! \vbl >{\columncolor[HTML]{EFEFEF}}P{0.9cm}|P{0.5cm}|P{0.8cm}! \vbl }

      \hbl
     \cellcolor[HTML]{C0C0C0}\textbf{Feature} & \cellcolor[HTML]{EFEFEF}\textbf{\#Cat} & \cellcolor[HTML]{EFEFEF}\textbf{\#Lang}  \\
      \hline
     144E & 23 & 38\\
      \hline
     144M & 23 & 45\\
      \hline
     144F & 22 & 48 \\
      \hline
     144O & 21 & 30\\
      \hbl

\end{tabular}}
\quad
{\renewcommand{\arraystretch}{\pad}
\begin{tabular}{! \vbl>{\columncolor[HTML]{EFEFEF}}P{0.9cm}|P{0.5cm}|P{0.8cm}! \vbl }

      \hbl
     \cellcolor[HTML]{C0C0C0}\textbf{Feature} & \cellcolor[HTML]{EFEFEF}\textbf{\#Cat} & \cellcolor[HTML]{EFEFEF}\textbf{\#Lang}  \\
      \hline
     83A & 0 & 1321\\
      \hline
     82A & 0 & 1302\\
      \hline
     97A & 0 & 1146\\
      \hline
     86A & 0 & 1083\\
      \hbl

\end{tabular}}

\caption{Most and least `ignored' typological features, the number of categories in each feature which have been ignored, and the number of languages which contain this feature.}
\label{tab:features}
 \end{center}
\end{table}

\begin{table}[!t]
\begin{center}
\small
{\renewcommand{\arraystretch}{\pad}
\begin{tabular}{! \vbl >{\columncolor[HTML]{EFEFEF}}P{1.2cm}|P{0.8cm}|P{1.3cm}|P{1.2cm}|P{0.9cm}! \vbl}

      \hbl
      \cellcolor[HTML]{C0C0C0}\textbf{Language} & \cellcolor[HTML]{EFEFEF}\textbf{Class} & \cellcolor[HTML]{EFEFEF}\textbf{\#Speakers} & \cellcolor[HTML]{EFEFEF}\textbf{`Ignored'} & \cellcolor[HTML]{EFEFEF}\textbf{Error}  \\
      \hline
     \LANG{Amharic} & 2 & 22M & 9 & 60.71 \\
      \hline
     \LANG{Breton} & 1 & 210k & 7 & 83.50 \\
      \hline
     \LANG{Swahili} & 2 & 18M & 8 & 45.64 \\
      \hline
     \LANG{Kabyle} & 1 & 5.6M & 8 & 39.10 \\
      \hbl

\end{tabular}}
\caption{Relevant examples of typologically `excluded' languages. The error rate is that of English $\rightarrow$ Language from \citet{massivemultilingualholger}.}
\label{tab:typology}
 \end{center}
\end{table}

\subsection{Findings}
\vskip 0.1cm
\textbf{Far-reaching repercussions} The most `ignored' feature in Table \ref{tab:features}, 144E (Multiple Negative Constructions in SVO Languages), is a rare feature, existing in only 38 languages over the world. These languages, however, are from various regions (e.g. \LANG{Wolof}, \LANG{Icelandic}, and \LANG{Kilivila}). Language tools in all these areas can be adversely affected without sufficient typological representation. On the other hand, common features such as 83A (Order of Object and Verb) are well represented with definite feature values for 1321 languages, ranging from \LANG{English} to \LANG{Mundari}.\vskip 0.2cm
\noindent
\textbf{Does it run in the family?}  \LANG{Amharic}, in Table \ref{tab:typology}, which among the Semitic family of languages, is the second most spoken language after \LANG{Arabic} (which has ~300M speakers). However, it has 9 `ignored' typological features, whereas \LANG{Arabic} has none. This reflects in the error rate of \LANG{English} to \LANG{Amharic} (60.71), which is significantly worse compared to 7.8 for \LANG{English} to \LANG{Arabic}.\\



%% file: sections/4_analysis.tex
\begin{figure*}[!t]
    \centering
        \begin{subfigure}{0.195\textwidth}
            \includegraphics[width=\columnwidth,keepaspectratio]{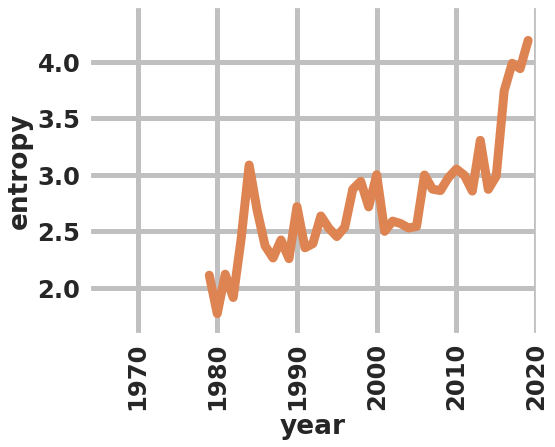}
            \caption{$c = $ \confacl}
        \end{subfigure}\hfill
        \begin{subfigure}{0.195\textwidth}
            \includegraphics[width=\columnwidth,keepaspectratio]{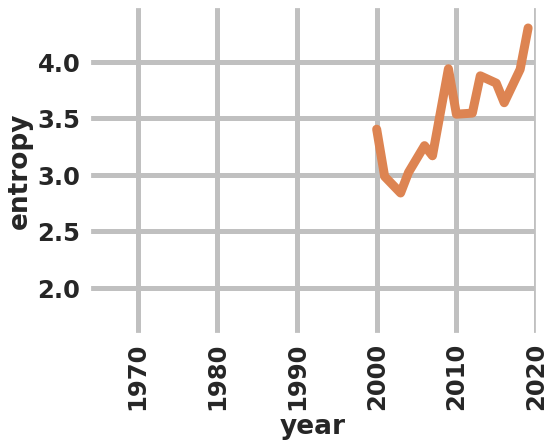}
            \caption{$c = $ \confnaacl}
        \end{subfigure}\hfill
        \begin{subfigure}{0.195\textwidth}
            \includegraphics[width=\columnwidth,keepaspectratio]{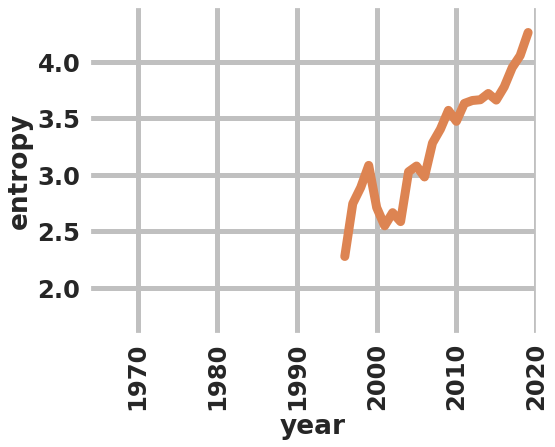}
            \caption{$c = $ \confemnlp}
        \end{subfigure}\hfill
        \begin{subfigure}{0.195\textwidth}
            \includegraphics[width=\columnwidth,keepaspectratio]{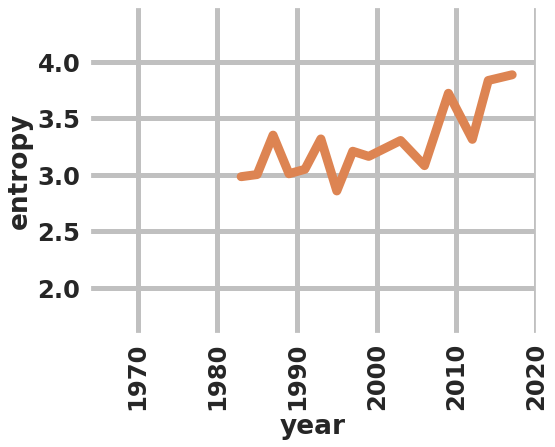}
            \caption{$c = $ \confeacl}
        \end{subfigure}\hfill
        \begin{subfigure}{0.195\textwidth}
            \includegraphics[width=\columnwidth,keepaspectratio]{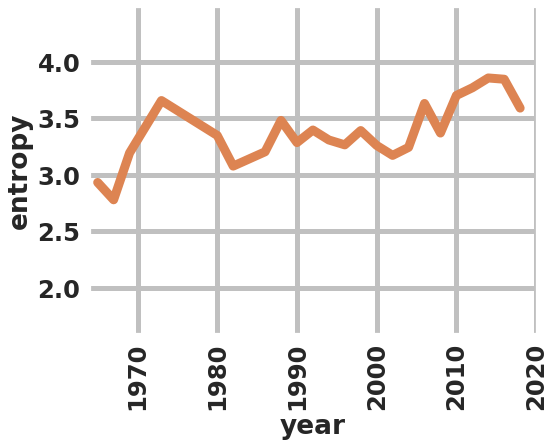}
            \caption{$c = $ \confcoling}
        \end{subfigure}

        \begin{subfigure}{0.195\textwidth}
            \includegraphics[width=\columnwidth,keepaspectratio]{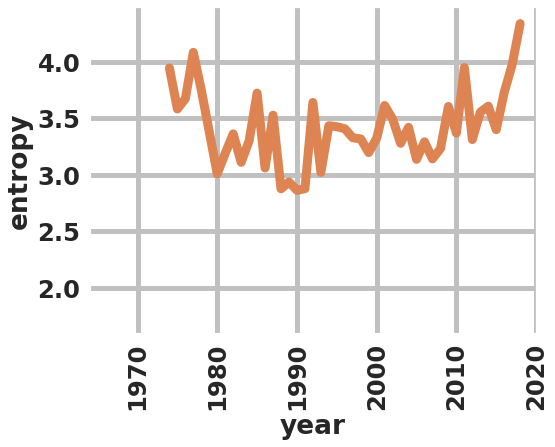}
            \caption{$c = $ \confcl}
        \end{subfigure}\hfill
        \begin{subfigure}{0.195\textwidth}
            \includegraphics[width=\columnwidth,keepaspectratio]{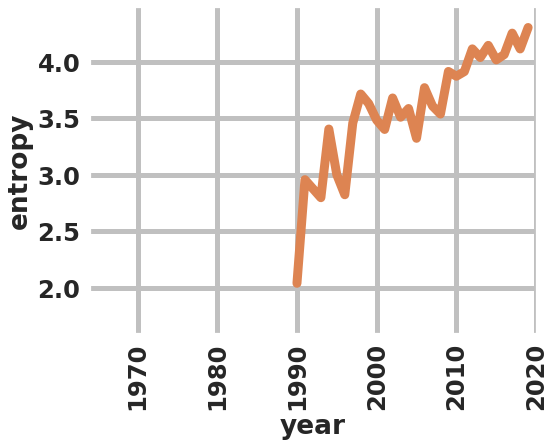}
            \caption{$c = $ \confws}
        \end{subfigure}\hfill
        \begin{subfigure}{0.195\textwidth}
            \includegraphics[width=\columnwidth,keepaspectratio]{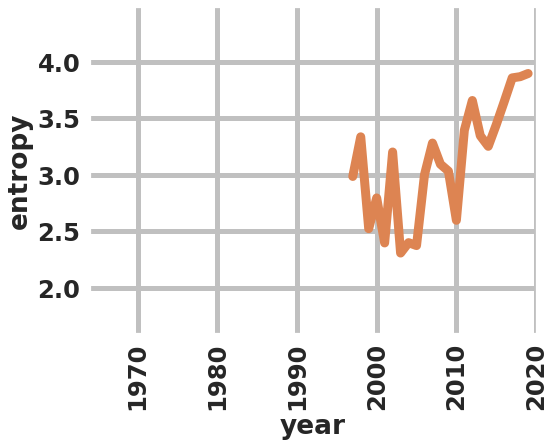}
            \caption{$c = $ \confconll}
        \end{subfigure}\hfill
      \begin{subfigure}{0.195\textwidth}
            \includegraphics[width=\columnwidth,keepaspectratio]{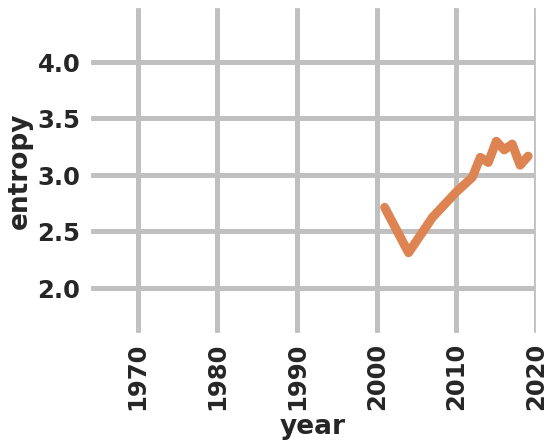}
            \caption{$c = $ \confsemeval}
        \end{subfigure}\hfill
        \begin{subfigure}{0.195\textwidth}
            \includegraphics[width=\columnwidth,keepaspectratio]{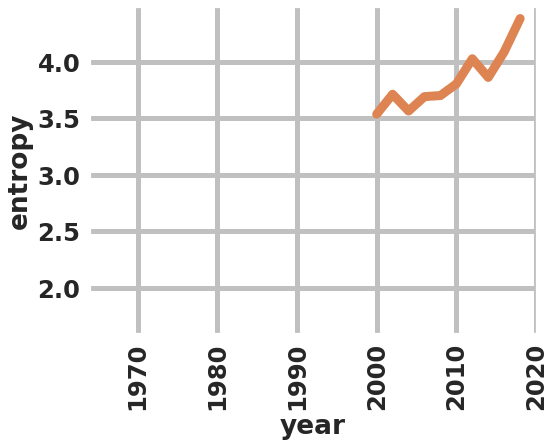}
            \caption{$c = $ \conflrec}
        \end{subfigure}
        
        \caption{Language occurrence entropy over the years for different conferences ($ \{ S\}^{c,y}$).}
        \label{fig:entropy}
\end{figure*}

\section{Conference-Language Inclusion}
NLP conferences have a huge impact on how language resources and technologies are constructed. Exciting research in venues such as \confacl, \confemnlp, \conflrec have the ability to turn heads in both industry and government and have the potential to attract funds to a particular technology. Has the usage of a small set of resource-rich languages in such conferences led to a disparity, pushing the less represented to the bottom of the ladder in terms of research? We analyze the involvement of various languages in NLP research conferences over the years.


\subsection{Dataset}

The ACL Anthology Corpus (ACL-ARC) \cite{bird2008acl} is the most extensively used dataset for analyzing trends in NLP research. This dataset contains PDFs, and parsed XMLs of Anthology papers. However, the latest versioned copy of ACL-ARC is till 2015 which makes it insufficient for analyzing trends in the most recent years. Moreover, paper data for non-ACL conferences such as \conflrec, \confcoling are absent from this dataset. In order to create a consistent data model, we augment this dataset by using Semantic Scholar's API and scraping ACL Anthology itself. Thus, we gather a consolidated dataset for 11 conferences which are relevant in judging global trends in NLP research. These include \confacl, \confnaacl, \confemnlp, \confeacl, \confcoling, \conflrec, \confconll, Workshops (\confws) (all since 1990), \confsemeval, \conftacl  and \confcl  Journals. We have attached the statistics of the dataset in Appendix \ref{sec:appendix}.

\subsection{Analysis}

\subsubsection{Language Occurrence Entropy}
The primary step of measuring the language diversity and inclusion of a conference and their progress is to measure the usage of language in that conference over multiple iterations. One of the ways to do it is by using frequency-based techniques where we can measure the occurrence of languages in that iteration. However, it is not a unified measure which represents the nature of language distribution with a single number. 
To this end, we use entropy as our metric to measure language inclusivity of each conference. It efficiently captures the skew in the distribution of languages, thereby making the disparity in language usage more clearer. The language occurrence entropy is calculated as follows:





For a conference $c$ held in year $y$ having $P$ papers, there exists a binary matrix $\{ M_{P \times L}\}^{c,y}$ where $M_{ij}$ is $1$ if $i^{th}$ paper ($\in P$) mentions the $j^{th}$ language ($\in L$). Then the entropy $\{ S\}^{c,y}$ is:
\begin{equation}
\begin{split}
    \{ S_{j}\}^{c,y} &= \frac{1}{P} \sum_{i=1}^{P} \{ M_{ij}\}^{c,y}\\
    \{ S'_{j}\}^{c,y} &= \frac{\{ S_{j}\}^{c,y}}{\sum_{j=1}^{L} \{ S_{j}\}^{c,y}} \\
    \{ S\}^{c,y} &= -\sum_{j = 1}^{L}\{ S'_{j}\}^{c,y}log_{e}\{ S'_{j}\}^{c,y}
\end{split}
\end{equation}
where $\{ S_{j}\}^{c,y}$ is a array of length $L$ accounting for number of papers in a specific language, $\{ S'_{j}\}^{c,y}$ is normalization done in order to get probability distribution for calculating entropy. In short, the higher the entropy, the more spread out is the distribution over the languages. The more peaked or skewed the distribution is, the lower is the entropy.

In Figure \ref{fig:entropy}, we can observe the entropy $S$ plotted for each $c$ as a function of $y$.

\subsubsection{Class-wise Mean Reciprocal Rank}
To quantify the extent of inclusion of language classes from our taxonomy in different conferences, we employ class-wise Mean Reciprocal Rank (MRR) as a metric. This helps in determining the standing of each class in a conference. If the rank of the language ($\textrm{rank}_{i}$) is ordered by the frequency of being mentioned in papers of a particular conference, and $Q$ is the total number of queries aka number of languages in each class, then:


\begin{equation}
    \textrm{MRR} = \frac{1}{|Q|} \sum_{i=1}^{|Q|} \frac{1}{\textrm{rank}_{i}}
\end{equation}

Table \ref{table:mrr} shows inverse mean reciprocal ranks of each category for a conference. The smaller the inverse MRR value, the more inclusive that conference is to that language class.
\begin{table}[!ht]
\small
\centering
{\renewcommand{\arraystretch}{\pad}

\begin{tabular}{! \vbl >{\columncolor[HTML]{EFEFEF}}P{1.63cm}|P{0.45cm}|P{0.45cm}|P{0.45cm}|P{0.45cm}|P{0.45cm}|P{0.45cm}! \vbl}
\hbl
\textbf{\scriptsize \cellcolor[HTML]{C0C0C0}Conf / Class} & \cellcolor[HTML]{EFEFEF}\textbf{0} & \cellcolor[HTML]{EFEFEF}\textbf{1} & \cellcolor[HTML]{EFEFEF}\textbf{2} & \cellcolor[HTML]{EFEFEF}\textbf{3} & \cellcolor[HTML]{EFEFEF}\textbf{4} & \cellcolor[HTML]{EFEFEF}\textbf{5} \\\hline
\confacl     & \zz{725} & \zz{372} & \zz{157} & \zz{63} & \zz{20} & \zz{3} \\\hline
\confcl      & \zz{647} & \zz{401} & \zz{175} & \zz{76} & \zz{27} & \zz{3} \\\hline 
\confcoling  & \zz{670} & \zz{462} & \zz{185} & \zz{74} & \zz{21} & \zz{2} \\\hline
\confconll   & \zz{836} & \zz{576} & \zz{224} & \zz{64} & \zz{16} & \zz{3} \\\hline
\confeacl    & \zz{839} & \zz{514} & \zz{195} & \zz{63} & \zz{15} & \zz{3} \\\hline
\confemnlp   & \zz{698} & \zz{367} & \zz{172} & \zz{67} & \zz{19} & \zz{3} \\\hline
\conflrec    & \zz{811} & \zz{261} & \zz{104} & \zz{45} & \zz{13} & \zz{2} \\\hline 
\confnaacl   & \zz{754} & \zz{365} & \zz{136} & \zz{63} & \zz{18} & \zz{3} \\\hline
\confsemeval & \zz{730} & \zz{ 983} & \zz{296} & \zz{121} & \zz{19} & \zz{3} \\\hline
\conftacl    & \zz{974} & \zz{400} & \zz{180} & \zz{50} & \zz{15} & \zz{3} \\\hline
\confws      & \zz{667} & \zz{293} & \zz{133} & \zz{59} & \zz{15} & \zz{3}
\\\hbl
\end{tabular}}
 \caption{Class-wise (1/MRR) for each conference.}
  \label{table:mrr}
\end{table}

\subsection{Findings}
\vskip 0.1cm
\textbf{All-Inclusive} Looking at the combined trends, both the entropy plots and the MRR figures suggest that \conflrec and \confws have been the most inclusive across all categories and have been continuing to do so over the years.\vskip 0.2cm
\noindent
\textbf{A ray of hope} With regards to the proceedings of \confacl, \confemnlp, \confnaacl, \conflrec, we note a marked spike in entropy in the 2010s, which is absent in other conferences. This might be due to the increased buzz surrounding cross-lingual techniques. \vskip 0.2cm
\noindent
\textbf{The later the merrier} An interesting point to note is that conferences which started later have taken lessons from past in matters of language inclusion. While the earlier established conferences have continued to maintain interest in a particular underlying theme of research which may or may not favour multilingual systems. This can be observed in : \confcoling, \confacl, \confeacl, \confemnlp (order of their start dates).\vskip 0.2cm
\noindent
\textbf{Falling off the radar} The taxonomical hierarchy is fairly evident when looking at the MRR table (Table \ref{table:mrr}) with class 5 coming within rank 2/3 and class 0 being `left-behind' with average ranks ranging from 600 to 1000. While the dip in ranks is more forgiving for conferences such as \conflrec, \confws, it is more stark in \confconll, \conftacl, \confsemeval.

%% file: sections/5_embeddings_v3.tex
\section{Entity Embedding Analysis}
The measures discussed in the previous section signal at variance in acceptance of different languages at different NLP venues across time. However, there are usually multiple subtle factors which vanilla statistics fail to capture. Embeddings, on the other hand, have been found extensively useful in NLP tasks as they are able to learn relevant signals directly from the data and uncover these rather complex nuances. To this end, we propose a novel approach to jointly learn the representations of \underline{conferences}, \underline{authors} and \underline{\smash{languages}}, which we collectively term as entities. The proposed embedding method allows us to project these entities in the same space enabling us to effectively reveal patterns revolving around them.



\subsection{Model}

We define the following model to jointly learn the embeddings of entities such that entities which have similar contextual distributions should co-occur together. For example, for an author {$A$}, who works more extensively on language {$L_{i}$} than {$L_{j}$} and publishes more at conference {$C_{m}$} than at conference {$C_{n}$}, the embeddings of { $A$} would be closer {$L_{i}$} than {$L_{j}$} and {$C_{m}$}  than {$C_{n}$}.

Given an entity and a paper associated with the entity, the learning task of the model is to predict $K$ randomly sampled words from the title and the abstract of the paper. We only select the title and abstract as compared to the entire paper text as they provide a concise signal with reduced noise. This model draws parallels to the Skipgram model of Word2Vec \cite{mikolov2013efficient}, where given an input word in Skipgram model, the task is to predict the context around the word. The input entity and $K$ randomly sampled words in our case correspond to the input word and context in the Skipgram model. The goal of the model is to maximize probability of predicting the random $K$ words, given the entity id as the input:

\begin{equation}
    \frac{1}{M}  \frac{1}{K} \sum_{m=1}^{M} \sum_{k=1}^{K} \sum_{i=1}^{I} \textit{p}(w_{k} | E_{<i,P_{j}>})
\end{equation}

where $E_{<i,P_{j}>}$ is the entity $E_{i}$ which is associated with the ${P_{j}}^{th}$ paper and $\textit{p}$ is the probability of predicting the word $w_{i}$ out of the $K$ words sampled from the paper and $M$ is the total number of papers in the dataset. To optimize for the above distribution, we define the typical SGD based learning strategy similar to Word2Vec\cite{mikolov2013efficient}. 

\begin{figure}
    \centering
    \includegraphics[width=0.48\textwidth]{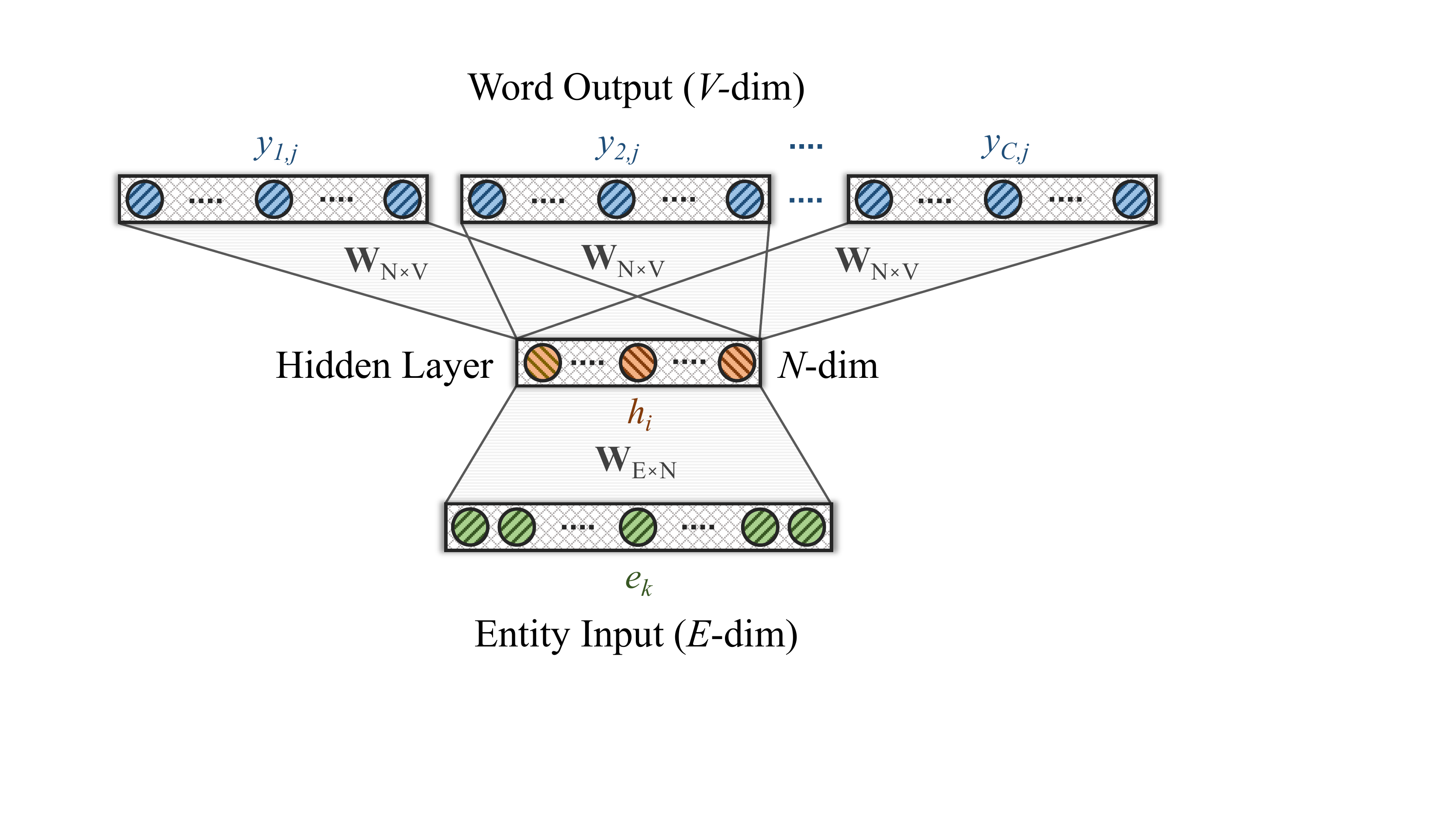}
    \caption{Model architecture to learn entity embeddings.  $W_{E \times N}$ is the weight matrix from input layer (entity layer) to the hidden layer, and $W_{N \times V}$ is the weight matrix for the hidden layer to output layer computation. At the end of training, $W_{E \times N}$ is the matrix containing embeddings of entities and $W_{N \times V}$ is the matrix containing the embeddings of words.}
    \label{fig:model}
\end{figure}

Figure \ref{fig:model} shows an outline of the model. The entity input layer has dimension equal to the total number of entities in the dataset ($E$). Hidden layer size is set to the desired embedding dimension ($N$). The output layer predicts words for the input entity and is of the same size as the vocabulary ($V$). The entities we learn are: (1) \underline{\smash{authors}} of the paper, (2) \underline{\smash{languages}} mentioned in the paper, (3) \underline{\smash{conference}} where the paper was accepted (e.g. \confacl), and (4) the \underline{\smash{conference iteration}} (e.g. ACL'19). We describe the model detail and hyperparameter tuning in Appendix \ref{sec:appendix}.  


\begin{table}[!ht]
\small
{\renewcommand{\arraystretch}{\pad}
\begin{tabular}{! \vbl >{\columncolor[HTML]{EFEFEF}}P{0.6cm}|P{1.2cm}|P{1.2cm}|P{1.2cm}|P{1.2cm}! \vbl}
\hbl
\cellcolor[HTML]{C0C0C0}\textbf{Class} & \cellcolor[HTML]{EFEFEF}\textbf{MRR(5)}  & \cellcolor[HTML]{EFEFEF}\textbf{MRR(10)} & \cellcolor[HTML]{EFEFEF}\textbf{MRR(15)} & \cellcolor[HTML]{EFEFEF}\textbf{MRR(20)} \\\hline
0     & \mrr{0.72281} & \mrr{0.69146} & \mrr{0.63852} & \mrr{0.57441} \\\hline
1     & \mrr{0.57210} & \mrr{0.52585} & \mrr{0.45354} & \mrr{0.40904} \\\hline
2     & \mrr{0.47039} & \mrr{0.45265} & \mrr{0.41521} & \mrr{0.38157} \\\hline
3     & \mrr{0.59838} & \mrr{0.52670} & \mrr{0.45131} & \mrr{0.42899} \\\hline
4     & \mrr{0.56016} & \mrr{0.47795} & \mrr{0.51199} & \mrr{0.50681} \\\hline
5     & \mrr{0.56548} & \mrr{0.51471} & \mrr{0.54326} & \mrr{0.47619}\\\hbl
\end{tabular}}
\caption{Language-Author-Language MRR on Taxonomy Classes. MRR({\bf K}) considers the closest {\bf K} authors.}
\label{MRR_Embeddings}
\end{table}



\begin{figure*}
    \centering
    \includegraphics[scale=0.4]{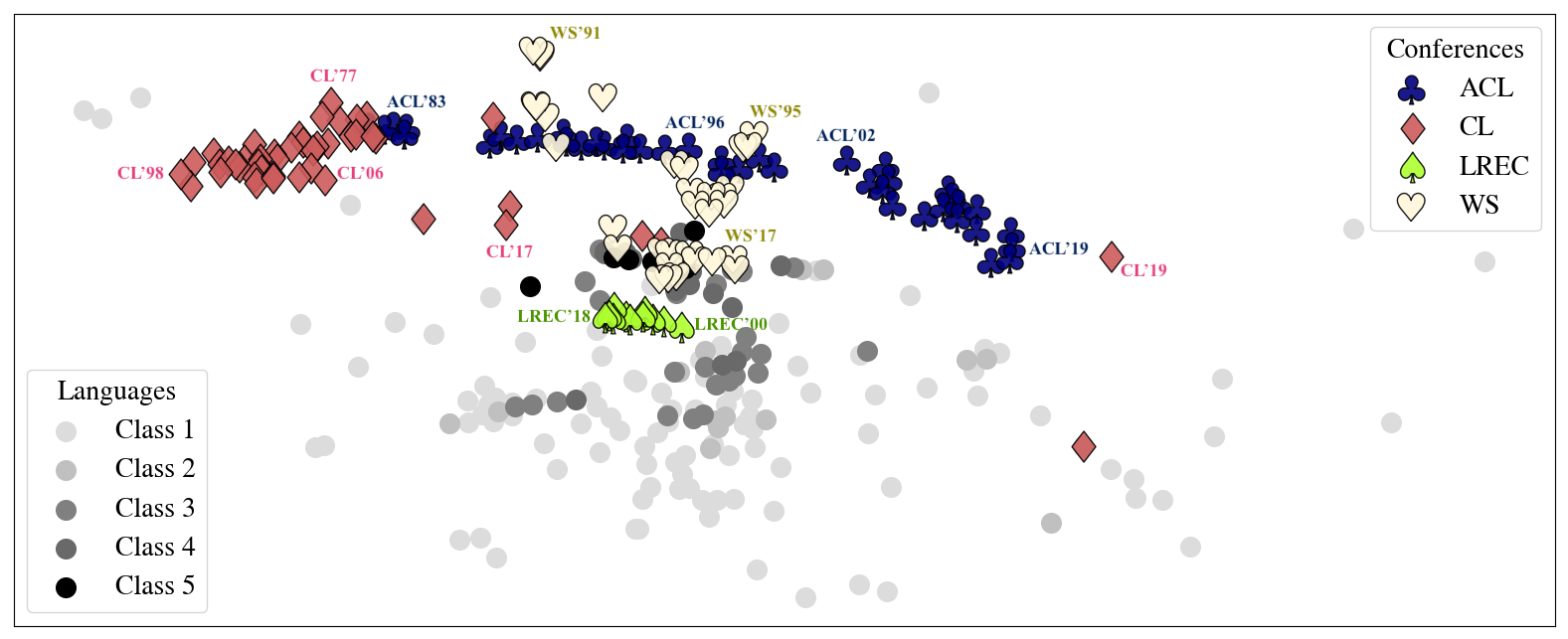}
    \caption{t-SNE visualization of the learnt conference and language embeddings.}
    \label{t-sne-conf}
\end{figure*}

\subsection{Analysis}
In order to better understand how languages are represented at different venues, we visualize the distribution of entity embeddings by projecting the generated embeddings into 2 dimensions using t-SNE \cite{maaten2008visualizing} (as shown in Figure \ref{t-sne-conf}). For clarity, we only plot \confacl, \conflrec, \confws and \confcl among the conferences, and all languages from the taxonomy, except those in Class 0. We omit plotting Class 0 languages as their projections are noisy and scattered due to their infrequent occurrence in papers.



To understand the research contributions of individual authors or communities towards research in respective language classes, we leverage the distribution between author and language entities by computing a variation of the Mean Reciprocal Rank (MRR). We consider a language {\bf L}, and take the {\bf K} closest authors to {\bf L} using cosine distance, and then take the closest {\bf M} languages to each author. If {\bf L} is present in the closest languages of an author, then we take the rank of {\bf L} in that list, inverse it, and average it for the {\bf K} authors. To compute this metric for a class of languages from the taxonomy, we take the mean of the MRR for all languages in that class. We fix {\bf M} to be 20, so as to understand the impact of the community when the number of languages remains unchanged. Table \ref{MRR_Embeddings} shows the MRR of various class of languages. A higher value of this measure indicates a more focused community working on that particular language, rather than a diverse range of authors.

\subsection{Findings}
\vskip 0.1cm
\textbf{Time waits for no conference} We can see a left to right trend in Figure \ref{t-sne-conf} with \confacl in 1983 in the left, and subsequent iterations laid out as we go right. We observe the same trend for \confeacl, \confnaacl, \confemnlp, \confconll, \conftacl, and \confcoling. We can say that the axis represents the progression of time to a certain extent. Alternatively, it may even represent a shift in the focus of NLP research, moving from theoretical research focused on grammar and formalisms on the left to a data-driven, more ML-oriented approach on the right. This can be observed as most of the \confcl embeddings are positioned on the left given their theoretical research focus.
\vskip 0.2cm
\noindent
\textbf{Long distance relationships?} From Figure \ref{t-sne-conf}, we can note that the less-resourced language classes are farther away from the trend-line of \confacl than the more resourced ones, with class 5 being closest, and class 1 being farthest. The visualization illustrates that languages are spreading out radially downwards from the \confacl trendline with popular classes of taxonomy like class 5 and class 4 being closer while others spreading out farther. Again, as previous analyses have shown us, \conflrec and \confws embeddings are closer to the language embeddings as compared to the other conferences as shown in Figure \ref{t-sne-conf}. In fact, \conflrec cluster is right in the middle of language clusters and so is the major part of the \confws cluster, especially in recent iterations.
\vskip 0.2cm
\noindent
\textbf{Not all heroes wear capes} 
Table \ref{MRR_Embeddings} shows the MRR for each class of languages in the taxonomy. From Table \ref{MRR_Embeddings}, it can be seen that class 0 has the highest MRR across different {\bf K} values. This shows that perhaps low resource languages have some research groups solely focused on the challenges related to them. There is a decreasing trend of MRR from class 0 to class 5, except for class 2, thereby indicating that more popular languages are addressed by more authors. We also observe that even though \LANG{Japanese}, \LANG{Mandarin}, \LANG{Turkish} and \LANG{Hindi} (MRR(10) $ >0.75$) are part of class 5 and class 4, their MRR is higher even compared to low resource languages in another classes, indicating that these languages have focused research communities working on them. On the other end of the spectrum, we observe a lot of low resource languages like \LANG{Burmese} (MRR(10) $ =0.02$), \LANG{Javanese} (MRR(10) $ =0.23$) and \LANG{Igbo} (MRR(10) $ =0.13$) which have millions of speakers but significantly low MRR values, potentially indicating that not a lot of attention is being given to them in the research community.

%% file: sections/7_conclusion.tex
\section{Conclusion}
\label{conclusion}
We set out to answer some critical questions about the state of language resource availability and research. We do so by conducting a series of quantitative analyses through the lens of a defined taxonomy. As a result, we uncover a set of interesting insights and also yield consistent findings about language disparity:
\vskip 0.2cm
\noindent
$\textbf{ --- }$ The taxonomical hierarchy is repeatedly evident from individual resource availabilities (LDC, LRE, Wikipedia, Web), entropy calculations for conferences, and the embeddings analysis.
\vskip 0.1cm
\noindent
$\textbf{ --- }$ \conflrec and Workshops(\confws) have been more inclusive across different classes of languages, seen through the inverse MRR statistics, entropy plots and the embeddings projection.
\vskip 0.1cm
\noindent
$\textbf{ --- }$ There are typological features (such as 144E), existing in languages over spread out regions, represented in many resource-poor languages but not sufficiently in resource-rich languages. This could potentially reduce the performance of language tools relying on transfer learning.
\vskip 0.1cm
\noindent
$\textbf{ --- }$ Newer conferences have been more language-inclusive, whereas older ones have maintained interests in certain themes of research which don't necessarily favour multilingual systems.
\vskip 0.1cm
\noindent
$\textbf{ --- }$ There is a possible indication of a time progression or even a technological shift in NLP, which can be visualized in the embeddings projection.
\vskip 0.1cm
\noindent
$\textbf{ --- }$ There is hope for low-resource languages, with MRR figures indicating that there are focused communities working on these languages and publishing works on them, but there are still plenty of languages, such as \LANG{Javanese} and \LANG{Igbo}, which do not have any such support.
\vskip 0.2cm

We believe these findings will play a strong role in making the community aware of the gap that needs to be filled before we can truly claim state-of-the-art technologies to be language agnostic. Pertinent questions should be posed to authors of future publications about whether their proposed language technologies extend to other languages. 

There are ways to improve the inclusivity of ACL conferences. Special tracks could be initiated for low-resource, language-specific tasks, although we believe that in doing so, we risk further marginalization of those languages. Instead, a way to promote change could be the addition of D\&I (Diversity and Inclusion) clauses involving language-related questions in the submission and reviewer forms: \textit{Do your methods and experiments apply (or scale) to a range of languages? Are your findings and contributions contributing to the inclusivity of various languages?}

Finally, in case you're still itching to know, Language {\bf X} is \LANG{Dutch}, and {\bf Y} is \LANG{Somali}.

\section{Acknowledgements}
We would like to thank Anshul Bawa, Adithya Pratapa, Ashish Sharma for their valuable feedback during the final phase of work. We would also like to thank the    anonymous reviewers for their many insightful comments and suggestions.

%% file: sections/8_appendix.tex










\newpage
\section{Appendix}
\label{sec:appendix}


\subsection{Embedding Visualization}
We have compiled a visualization of the embedding space of conferences and languages which can be run on a browser. This is available as an interactive visualization on \href{https://microsoft.github.io/linguisticdiversity}{https://microsoft.github.io/linguisticdiversity}, and can be used to play around with different combinations to see how NLP research has progressed over the years in terms of language inclusion. The legends are self-explanatory and are clickable to add or remove those points. The numbers in the legend represent the respective classes.

\subsection{ACL Anthology Dataset Statistics}
We have accounted for all the papers which have appeared in the main track proceedings of the conference. This includes all the long and short papers and excludes System Demonstrations, Tutorial Abstracts, Student Research Workshops, Special Issues, and other such tracks out of the scope of measuring language usage trends in general NLP research. We are in the process of releasing the dataset along with the documentation.

\begin{table}[!h]
\small
{\renewcommand{\arraystretch}{\pad}
\begin{tabular}{! \vbl >{\columncolor[HTML]{EFEFEF}}l|l|l|l|l! \vbl }
\hbl
\cellcolor[HTML]{C0C0C0} \textbf{\scriptsize Conf / Class}        & \cellcolor[HTML]{EFEFEF}\textbf{\scriptsize \#Papers} & \cellcolor[HTML]{EFEFEF}\textbf{\scriptsize \#Body} & \cellcolor[HTML]{EFEFEF}\textbf{\scriptsize \#NoProc} & \cellcolor[HTML]{EFEFEF}\textbf{\scriptsize \% Missing} \\ \hline
\textbf{\conflrec}      & 5835 & 15 & 6 & 0.1\% \\ \hline
\textbf{\confws}        & 17844 & 337 & 332 & 1.86\% \\ \hline
\textbf{\confconll}     & 1035 & 0 & 0 & 0.0\% \\ \hline
\textbf{\confeacl}      & 1165 & 4 & 1 & 0.09\% \\ \hline
\textbf{\confacl}       & 5776 & 46 & 29 & 0.5\% \\ \hline
\textbf{\conftacl}      & 280 & 7 & 0 & 0.0\% \\ \hline
\textbf{\confcl}        & 2025 & 88 & 0 & 0.0\% \\ \hline
\textbf{\confnaacl}     & 2188 & 2 & 1 & 0.05\% \\ \hline
\textbf{\confcoling}    & 4233 & 5 & 2 & 0.05\% \\ \hline
\textbf{\confemnlp}     & 3865 & 16 & 16 & 0.41\% \\ 
\hbl
\end{tabular}}
\caption{Dataset Statistics.}
\label{tab:datastats}
\end{table}

\subsection{Hyperparameter Tuning}
Our model has same hyperparameters as that of Word2Vec. To determine the optimal hyper-parameters for  the model, we take the entire dataset and split it into a 80-20 ratio, and given the embedding of a paper, the task is to predict the year in which the paper is published. Given this vector for a paper, we use a linear regression model such that given this vector, the model is supposed to predict the year in which the paper was published. We measured both $R^2$ measure of variance in regression and mean absolute error (MAE). $R^2$ is usually in the range of 0 to 1.00 (or 0 to 100\%) where 1.00 is considered to be the best. MAE has no upper bound but the smaller it is the better, and 0 is its ideal value. We observed that our model does not show significant difference across any hyperparaeters except for the size of embeddings. The best dimension size for our embeddings is 75, and, we observed the corresponding $R^2$ value of 0.6 and an MAE value of 4.04.

\subsection{Cosine distance between conferences and languages}
From Figure \ref{t-sne-conf}, we can see that languages are somewhat below the conferences are closer to some conferences while distant from others. To quantify this analysis, we compute the cosine distance between the conference vector and the mean of the vector each category of the taxonomy. Table \ref{distance_vectors} shows the cosine distance between the conferences and the each category of languages and we see a very similar trend that while \confacl is an at average distance of 0.291 from category 5 languages, its almost more than double far away from category 2. There is also a very steep rise in distance of the \confacl vector from category 4 to category 3. In fact, similar trends are visible for other \confacl related conferences including \confeacl, \confnaacl, \confemnlp and \conftacl. We can also see that in Table \ref{distance_vectors}, \confws and \conflrec are closest from category 2 to category 5 whereas almost all conferences are somewhat at the same distance from category, except the CL journal. The trend for category 0 languages seems somewhat different than the usual trend is this table, probably because of the large number of languages in this category as well as the sparsity in papers.
\begin{table}[!h]
\small
{\renewcommand{\arraystretch}{\pad}
\begin{tabular}{! \vbl >{\columncolor[HTML]{EFEFEF}}l|l|l|l|l|l|l! \vbl}
\hbl
\cellcolor[HTML]{C0C0C0} \textbf{\scriptsize Conf / Class}        & \cellcolor[HTML]{EFEFEF}\textbf{0} & \cellcolor[HTML]{EFEFEF}\textbf{1} & \cellcolor[HTML]{EFEFEF}\textbf{2} & \cellcolor[HTML]{EFEFEF}\textbf{3} & \cellcolor[HTML]{EFEFEF}\textbf{4} & \cellcolor[HTML]{EFEFEF}\textbf{5} \\ \hline
\textbf{\conflrec}      & 0.51    & 0.51    & 0.52     & 0.42    & 0.36    & 0.32    \\ \hline
\textbf{\confws} & 0.50    & 0.55    & 0.53    & 0.40    & 0.28    & 0.21    \\ \hline
\textbf{\confconll}     & 0.54    & 0.60    & 0.63    & 0.49     & 0.40    & 0.46    \\ \hline
\textbf{\confeacl}      & 0.53    & 0.55     & 0.59     & 0.45    & 0.34    & 0.32     \\ \hline
\textbf{\confacl}       & 0.48    & 0.51   & 0.60    & 0.42     & 0.34    & 0.29    \\ \hline
\textbf{\conftacl}      & 0.52    & 0.56    & 0.66    & 0.48    & 0.38    & 0.47    \\ \hline
\textbf{\confcl}        & 0.67    & 0.78    & 0.80    & 0.75    & 0.65    & 0.59    \\ \hline
\textbf{\confnaacl}     & 0.48    & 0.52    & 0.59    & 0.47    & 0.39    & 0.33    \\ \hline
\textbf{\confcoling}    & 0.48    & 0.53    & 0.55    & 0.46    & 0.37    & 0.30    \\ \hline
\textbf{\confemnlp}     & 0.57    & 0.59    & 0.66    & 0.51    & 0.46    & 0.45   \\ \hbl
\end{tabular}}
\caption{Cosine Distance between conference vectors and mean class vectors of languages.}
\label{distance_vectors}
\end{table}


\subsection{Taxonomy classification}
We release our full language taxonomy classification on the website: \href{https://microsoft.github.io/linguisticdiversity}{https://microsoft.github.io/linguisticdiversity}.

\subsection{Class-wise log(MRR) over the years per conference}
We plot MRR on a log scale for each conference to measure the progress of inclusion of the defined taxonomy classes over the years. It is very interesting to note how \conflrec has very smooth forward progression.
\begin{figure}[!ht]
    \centering
    \begin{subfigure}{0.46\textwidth}
            \includegraphics[width=\columnwidth,keepaspectratio]{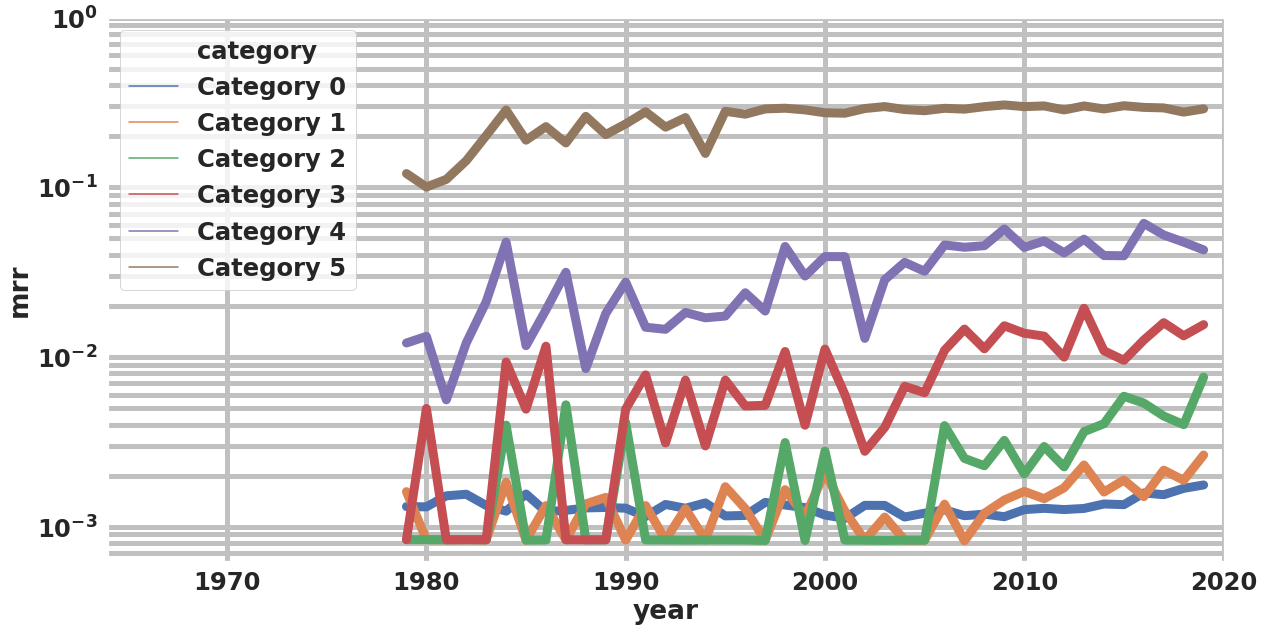}
            \caption{$c = $ \confacl}
    \end{subfigure}
    \begin{subfigure}{0.46\textwidth}
            \includegraphics[width=\columnwidth,keepaspectratio]{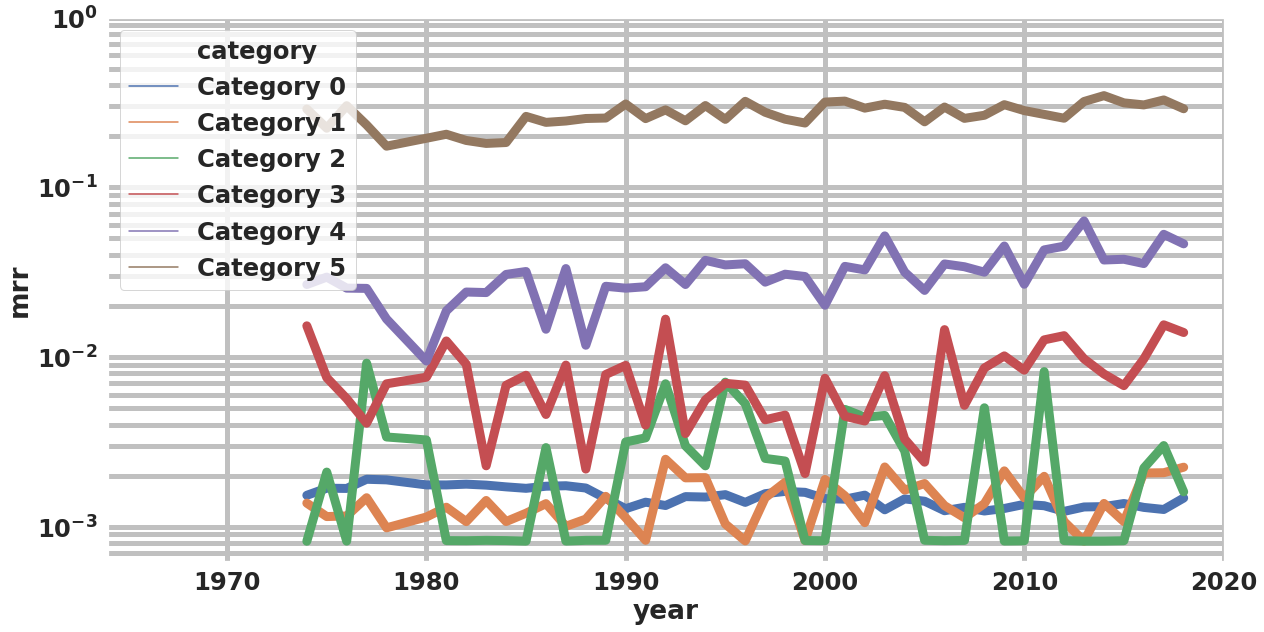}
            \caption{$c = $ \confcl}
    \end{subfigure}
    \begin{subfigure}{0.46\textwidth}
            \includegraphics[width=\columnwidth,keepaspectratio]{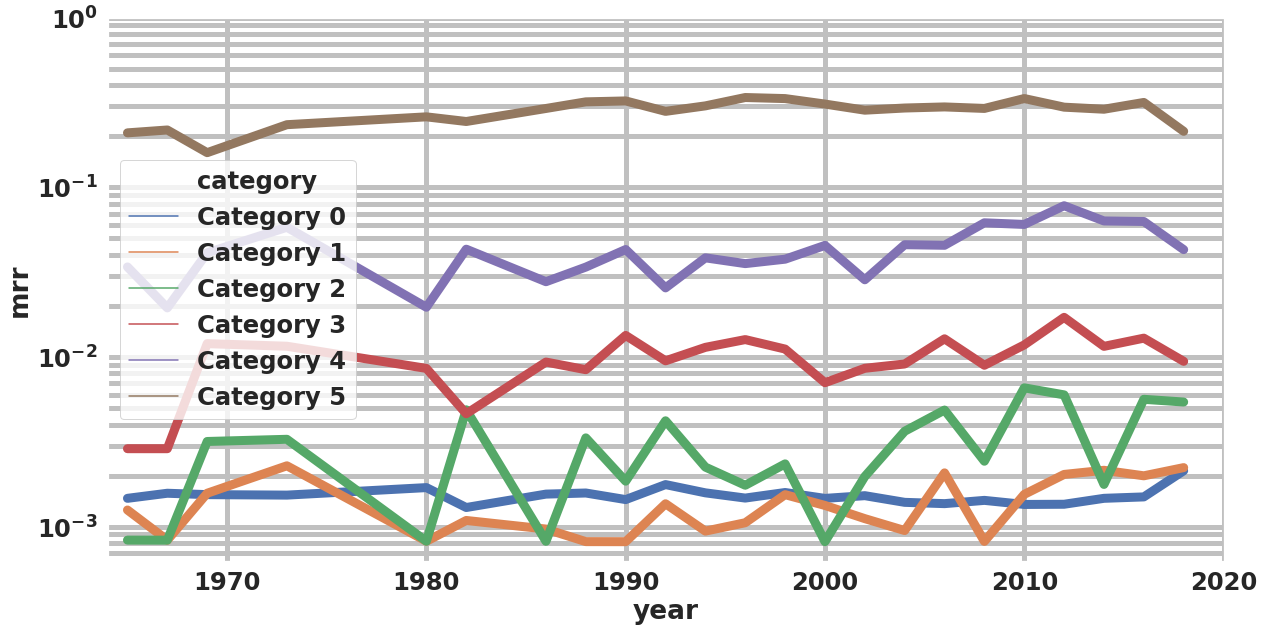}
            \caption{$c = $ \confcoling}
    \end{subfigure}
    \begin{subfigure}{0.46\textwidth}
            \includegraphics[width=\columnwidth,keepaspectratio]{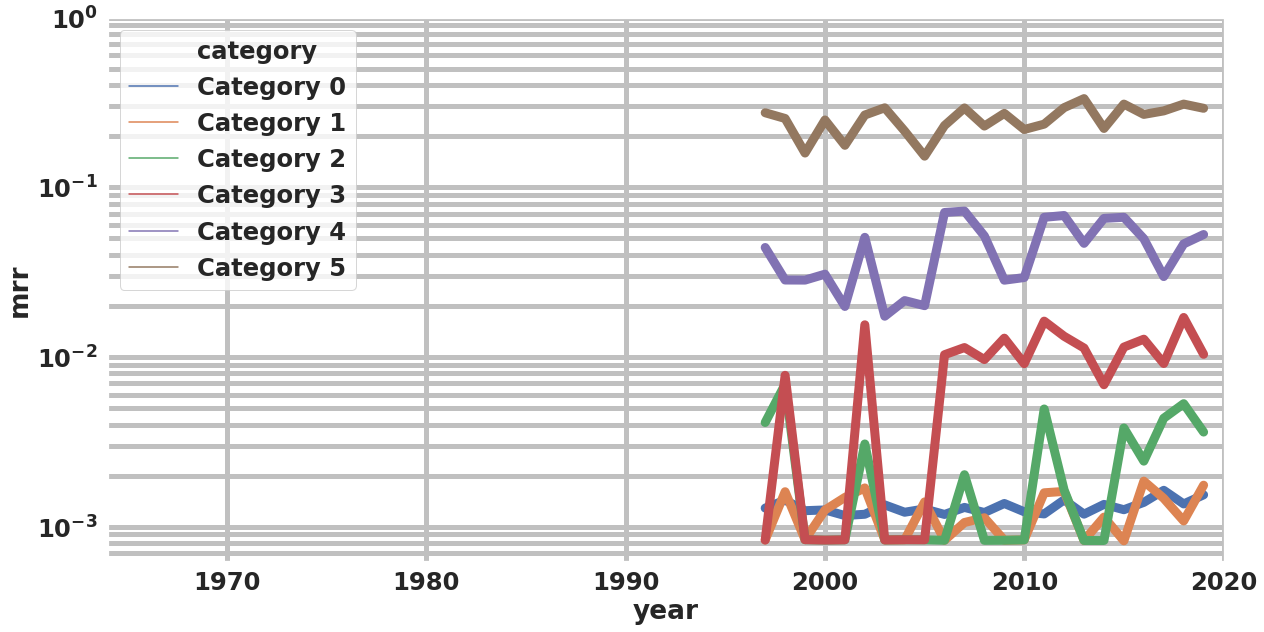}
            \caption{$c = $ \confconll}
    \end{subfigure}
\end{figure}

\begin{figure}[!ht]
\vskip 1cm
\begin{subfigure}{0.46\textwidth}
            \includegraphics[width=\columnwidth,keepaspectratio]{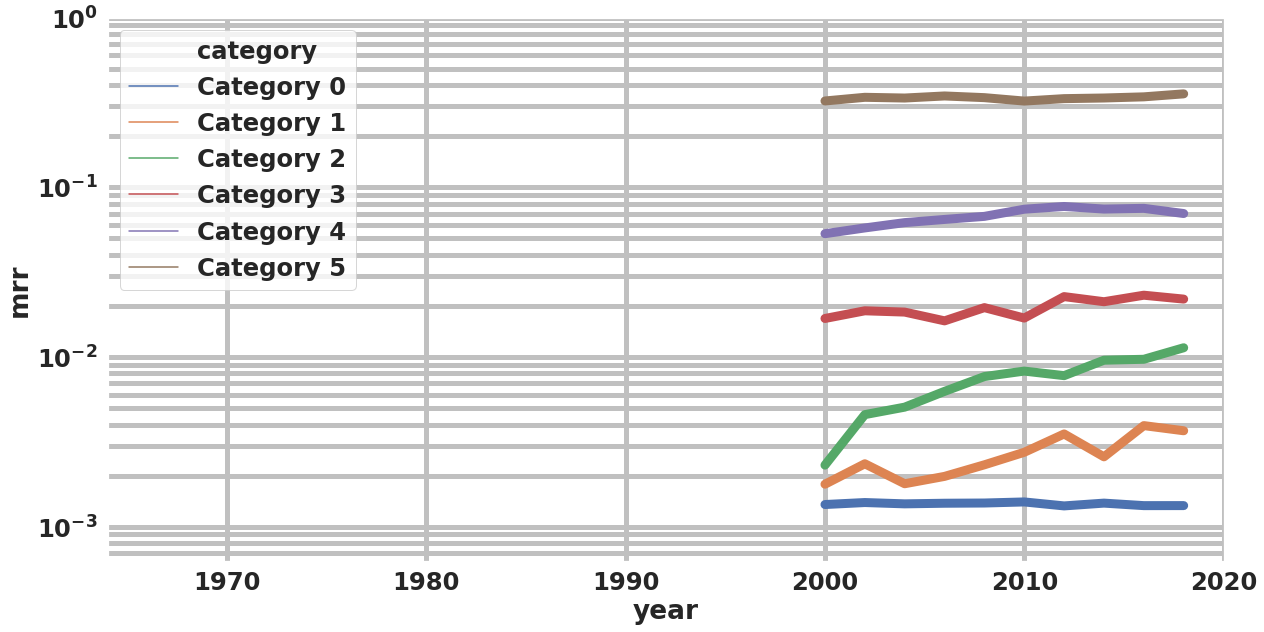}
            \caption{$c = $ \conflrec}
        \end{subfigure}
    \begin{subfigure}{0.46\textwidth}
            \includegraphics[width=\columnwidth,keepaspectratio]{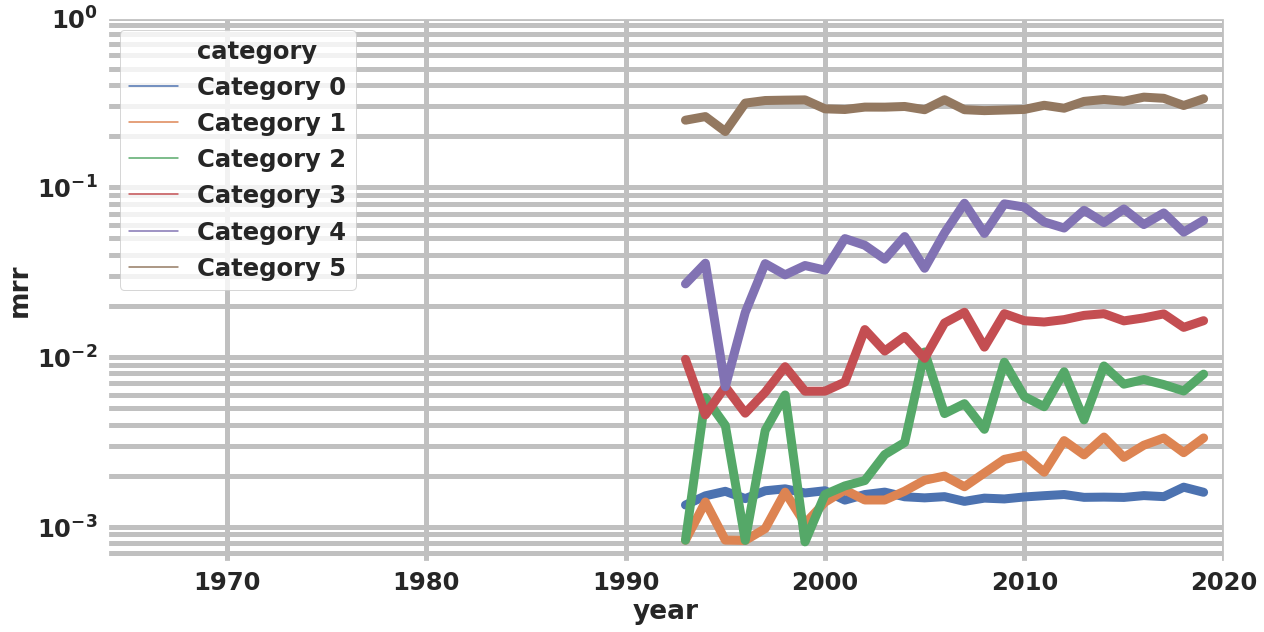}
            \caption{$c = $ \confws}
        \end{subfigure}
\begin{subfigure}{0.46\textwidth}
            \includegraphics[width=\columnwidth,keepaspectratio]{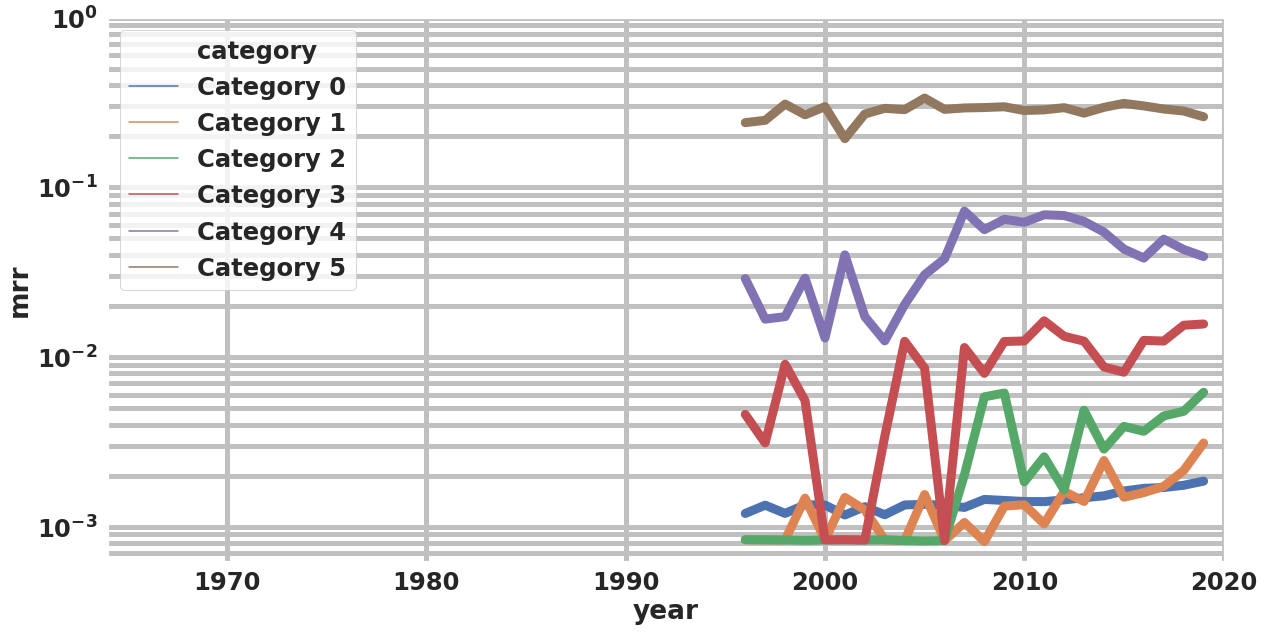}
            \caption{$c = $ \confemnlp}
        \end{subfigure}
\begin{subfigure}{0.46\textwidth}
            \includegraphics[width=\columnwidth,keepaspectratio]{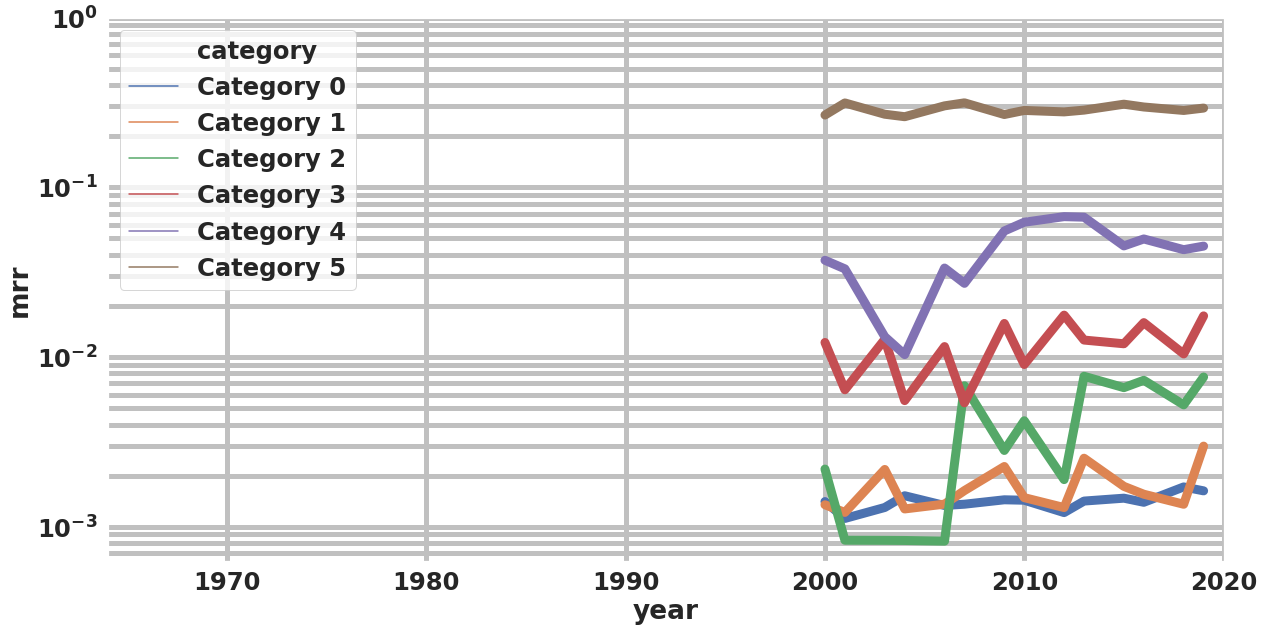}
            \caption{$c = $ \confnaacl}
        \end{subfigure}
\begin{subfigure}{0.46\textwidth}
            \includegraphics[width=\columnwidth,keepaspectratio]{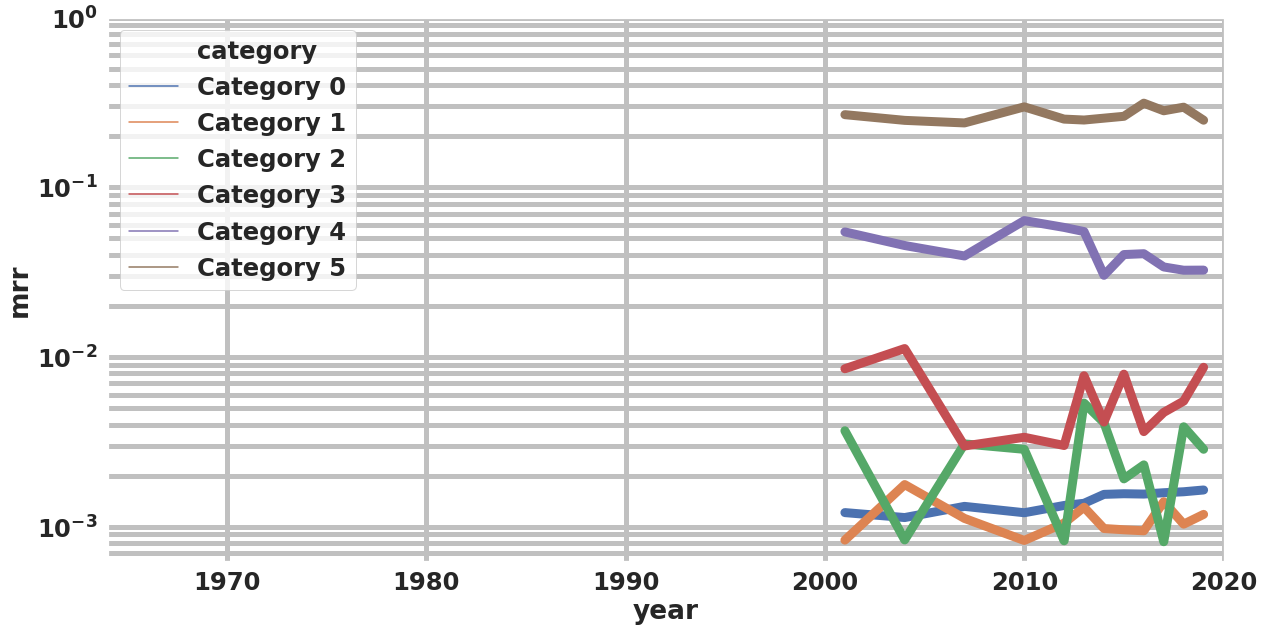}
            \caption{$c = $ \confsemeval}
        \end{subfigure}
\end{figure}